\documentclass[10pt,twocolumn,letterpaper]{article}

\usepackage{cvpr}              

\usepackage[utf8]{inputenc}
\usepackage{textcomp}
\usepackage{algorithm}
\usepackage{newunicodechar}
\newunicodechar{ }{\,}

\definecolor{cvprblue}{rgb}{0.21,0.49,0.74}
\usepackage[pagebackref,breaklinks,colorlinks,allcolors=cvprblue]{hyperref}

\def\paperID{44973}
\def\confName{CVPR}
\def\confYear{2026}

\title{Pixelis: Reasoning in Pixels, from Seeing to Acting}

\author{
    Yunpeng Zhou \\
    University of Reading, Reading, UK \\
    \texttt{kc804139@student.reading.ac.uk}
}

\usepackage{makecell}
\usepackage{booktabs}
\usepackage{caption}
\usepackage{amsmath}
\usepackage{tabularx}
\usepackage{amsmath,amssymb}
\usepackage{graphicx}
\usepackage{booktabs}
\usepackage{multirow}
\usepackage{algorithm}
\usepackage{algpseudocode}
\floatname{algorithm}{Algorithm}
\usepackage{newtxtext,newtxmath}
\usepackage{microtype}
\usepackage{xurl}
\usepackage{enumitem}
\usepackage{booktabs,array,ragged2e}
\usepackage{adjustbox}
\algrenewcommand\algorithmicrequire{\textbf{Input:}}
\algrenewcommand\algorithmicensure{\textbf{Output:}}

\begin{document}
\maketitle
\begin{abstract}
Most vision-language systems are static observers: they describe pixels, do not act, and cannot safely improve under shift. This passivity limits generalizable, physically grounded visual intelligence. Learning through action, not static description, is essential beyond curated data. We present Pixelis, a pixel-space agent that operates directly on images and videos via a compact set of executable operations (zoom/crop, segment, track, OCR, temporal localization) and learns from its consequences. Pixelis trains in three phases: (1) Supervised Fine-Tuning learns a pixel-tool grammar from Chain-of-Thought-Action traces with a masked imitation loss that upweights operation/argument tokens and auxiliary heads to stabilize pixel-grounded arguments; (2) Curiosity-Coherence Reward Fine-Tuning optimizes a dual-drive objective marrying prediction-error curiosity with adjacent-step coherence and a mild efficiency prior under a KL anchor, yielding short, valid, structured toolchains; (3) Pixel Test-Time RL performs label-free adaptation by retrieving neighbors, voting over complete trajectories rather than answers, and updating toward short, high-fidelity exemplars while constraining drift with a KL-to-EMA safety control. Across six public image and video benchmarks, Pixelis yields consistent improvements: the average relative gain is +4.08\% over the same 8B baseline (peaking at +6.03\% on VSI-Bench), computed as (ours-baseline)/baseline, while producing shorter, auditable toolchains and maintaining in-corridor KL during test-time learning. Acting within pixels, rather than abstract tokens, grounds multimodal perception in the physical world, linking visual reasoning with actionable outcomes, and enables embodied adaptation without external feedback.
\end{abstract}
    
\begin{figure*}[t]
  \centering
  \includegraphics[width=0.88\linewidth]{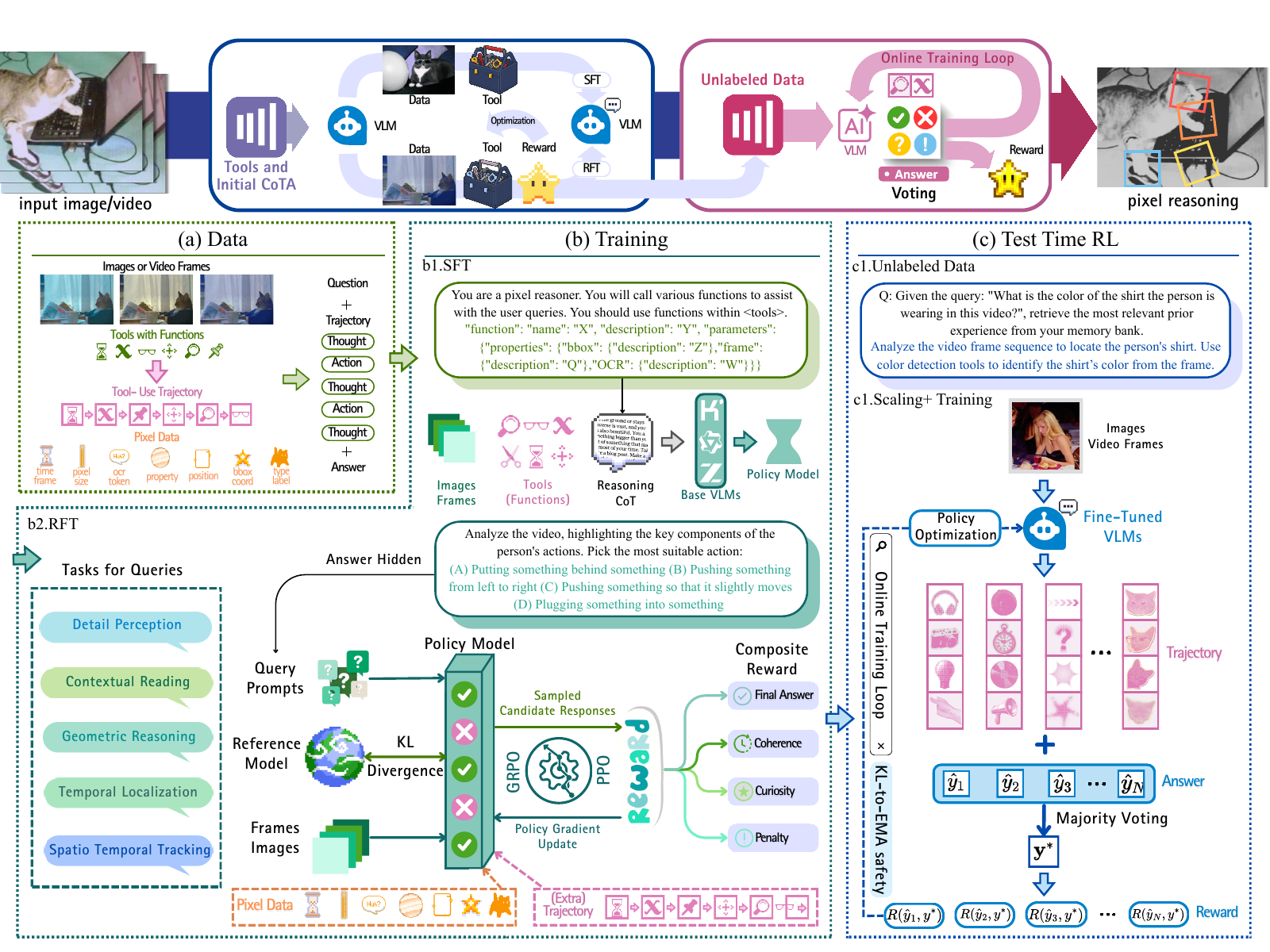}
  \vspace{-8pt}
  \caption{Pixelis uses executable pixel tools to act on images and videos. SFT learns tool syntax; CC-RFT shapes exploration with curiosity, coherence, and efficiency; Pixel TTRL adapts online via retrieval and trajectory voting under KL/EMA safety, producing shorter, auditable toolchains.}
  \label{fig:your_label}
\end{figure*}

\section{Introduction}
\label{sec:intro}
Large multimodal models excel at describing pixels yet remain passive observers. They neither act on visual evidence nor learn from their interactions, breaking the perception–action loop and allowing localization, reading, and timing errors to persist under domain shift. This work introduces pixel-space agents that execute verifiable tools and improve safely at test time, closing this loop by grounding each reasoning step in direct visual evidence.

We present Pixelis, a vision–language agent that composes auditable toolchains and refines them online. Its three-stage pipeline begins with Supervised Fine-Tuning (SFT) to learn tool-use syntax from Chain-of-Thought–Action traces. Next, Curiosity–Coherence RFT (CC-RFT) balances visual exploration with an adjacent-step coherence prior that discourages erratic tool switching without stifling longer-horizon reasoning. Finally, Pixel TTRL enables safe online adaptation by voting over full reasoning trajectories: rather than tallying only final answers, it selects behaviorally consistent exemplars to prevent drift from accumulated errors.

Reasoning in pixel space raises three challenges. Executability: map decisions to typed, replayable tool calls. Structure: prefer concise, logically ordered steps over chaotic exploration. Safety: ensure stable, auditable updates under shift. Pixelis addresses each directly. Executability comes from a compact tool interface with serialized arguments. Structure is induced by coherence on step embeddings that favors locally adjacent effects and discourages gratuitous detours. Safety is provided by turning high-confidence trajectories into supervision, while a KL-to-EMA corridor acts as a policy compass, penalizing updates that veer too far from a slow, stable reference so refinement stays purposeful rather than oscillatory.

These elements reinforce one another. Adjacent-step coherence shortens plans, which makes trajectory-level voting more reliable; in turn, Pixel TTRL can update aggressively while keeping token-level KL inside a corridor. At 8B scale across six benchmarks, Pixelis improves accuracy while shortening chains from $\approx$ 6 to 3.7 steps on average and keeping token-KL $<$0.2; removing safety pushes KL $>$0.4 with accuracy collapse. To attribute gains to actions rather than text-only heuristics, we report tool fidelity (IoU/ANLS/HOTA), include tool-ablated controls and tool-dependent subsets, and audit de-duplication and retrieval leakage.

Our evaluation is process-oriented. We introduce RaPR/RaCPR to measure valid tool use and coherent composition, analyze risk–coverage for adaptation with matched budgets, and separate pre/post scores under shift. Adversarial-majority stress tests show that uncertainty-weighted consensus with abstention keeps KL in-corridor and improves accuracy where hard majority drifts. A de-duplication and leakage audit confirms index whitelisting and low residual overlap, supporting fair comparison to static baselines.

Contributions: Pixelis unifies pixel-space acting with a small set of executable operations composed into short, replayable, verifiable toolchains that reveal how answers are obtained; introduces CC-RFT, coupling prediction-error curiosity with adjacent-step coherence and an efficiency prior to avoid noise-seeking while keeping plans compact; and proposes Pixel TTRL, a trajectory-level behavioral voting scheme that supplies reliable test-time supervision without a process critic, safeguarded by KL-to-EMA drift control. Together these components deliver consistent 8B-scale gains across six benchmarks while producing shorter chains, maintaining in-corridor KL during online adaptation, and clarifying when and why pixel tools—rather than text-only heuristics—drive the improvements, with released configs, checkpoints, replay logs, and scripts enabling end-to-end replication, ablations, and audited updates.

\section{Related Work}
\label{sec:related_work}
Our work connects tool-augmented VLMs, RL for structured visual reasoning, and online test-time adaptation. Prior systems typically excel in subsets of this space—pixel grounding, structured exploration, or safe updates—yet rarely unify all three. We pursue this combination via a compact, auditable 
pixel toolchain with curiosity–coherence training and KL-anchored trajectory voting, enabling verifiable stepwise execution and explicit chain-length control.

\noindent\textbf{Tool-augmented VLMs.}
External tools expand model capability and tool-grounded intermediates~\cite{schick2023toolformer,ma-etal-2024-sciagent,davis2025augmentedvisionlanguagemodelssystematic}. Early agents (Visual ChatGPT, MM-REACT) orchestrate vision APIs with limited stepwise verification~\cite{wu2023visualchatgpt,yang2023mmreact}. Visual Sketchpad makes intermediate reasoning visible through sketch editing~\cite{chen2024visualsketchpad}. PixelLM performs internal mask generation via a codebook, removing external APIs but not avoiding execution~\cite{ren2024pixellm}. PixelWorld (arXiv 2025) standardizes pixel-level evaluation but does not prescribe tool design~\cite{lyu2025pixelworld}. AnyTool scales API retrieval yet omits test-time behavioral consistency~\cite{du2024anytool}. Recent work in autonomous driving explores tool-augmented VLM agents and vision–language–action models: AgentThink unifies chain-of-thought with dynamic tool invocation for driving scenarios~\cite{qian2025agentthink}, while surveys and frameworks for vision–language–action driving agents and interactive traffic generation~\cite{jiang2025vla_survey} emphasize structured perception–action interfaces rather than pixel-audited toolchains. Our setting differs by targeting short, auditable pixel chains aligned with VTG-style localization~\cite{wu2025surveyvideotemporalgrounding} and furnishing replayable execution traces.

\noindent\textbf{RL for structured reasoning.}
RLHF and sequence-level RL enable trajectory learning, and recent reason-tuning couples RL with chain supervision~\cite{christiano2017deep,ouyang2022training,tan2025reason}. 
Self-consistency~\cite{wang2022selfconsistency} and constitutional variants rely on 
costly process signals~\cite{lightman2023letsverify,bai2022constitutional,yuan2023selfrewarding}. 
Intrinsic motivation encourages exploration~\cite{pathak2017curiosity}. 
Step-level coherence has been studied in language models 
and robotics~\cite{osa2018motionprimitives}, but applying adjacent-step coherence 
to pixel-space tool chains remains underexplored. CC-RFT introduces via z-scored cosine similarity on step embeddings, building structured plans 
without per-step critics.

\noindent\textbf{Test-time adaptation.}
Entropy-minimization (TENT), TTS (Test Time Scaling) and continual TTA (CoTTA, RoTTA) update without pixel evidence and can drift~\cite{wang2021tent,wang2022cotta,liu2022rotta,snell2024scalingllmtesttimecompute,zhang2025surveytesttimescalinglarge}. For VLMs, online and adapter-based TTA show similar stability–performance issues~\cite{döbler2024lostopportunityvisionlanguagemodels,karmanov2024efficienttesttimeadaptationvisionlanguage}. TTRL casts this as test-time reinforcement learning using voting-derived pseudo-rewards~\cite{zuo2025ttrl}. KL penalties are standard in RL and TTA~\cite{schulman2017proximal,Liang_2024,wang2024searchlostonlinetesttime}; we adapt them via a slow-moving EMA anchor to stabilize non-stationary updates. VIMA~\cite{jiang2023vima} 
couples perception and action in robot control, whereas Pixelis targets semantic VLM reasoning with verifiable tool outputs. All comparisons use backbone-matched controls and the same KL corridor.

\section{Method}
\vspace{-2pt}
\subsection{Data}
\label{sec:data}

Pixelis trains on Chain-of-Thought–Action (CoTA) trajectories that log how pixel tools are chosen and updated.

\noindent\textbf{Scope and collection.}
We build a corpus of CoTA trajectories interleaving reasoning with executable pixel tools. Each step selects a tool \{SEG, TRK, OCR, ZOOM, TEMP, PROP\}, sets arguments, and records the new state. Trajectories come from a tool-constrained teacher using the Grok 4 VLM (API, 2025 Oct, temp.\ 0.7) ~\cite{xai2025grok4}with logit-biased syntax. To reduce tool bias, we stratify by task, tool, and temporal span and schedule difficulty via a small SFT hardness predictor. We drop duplicates, license conflicts, and visible watermarks.

\noindent\textbf{Validation and acceptance.}
Operator checks include IoU (SEG/TRK), edit distance (OCR), and temporal consistency (TEMP). We compute a trajectory score
\[
\begin{aligned}
S(\tau) &= \alpha S_{\text{logic}}
          + \beta S_{\text{struct}}
          + \gamma S_{\text{visual}},\\[-2pt]
\tau \in \mathcal{D} &\iff (\text{all ops pass}) \wedge (S(\tau) \ge \tau_0).
\end{aligned}
\]
with dev-tuned defaults $\alpha{=}0.4$, $\beta{=}0.3$, $\gamma{=}0.3$, $\tau_0{=}0.65$; components are z-scored per task to balance magnitudes. A decisive step is a succeeded call that changes the intermediate state or final answer; decisive-step rates are computed over succeeded calls per split.

\noindent\textbf{Core statistics.}
Images (train/dev/test): 80k/8k/8k; Videos: 28k/2k/2k.
CoTA trajectories: 76k with 5.8 steps on average, longer than online chains.
Decisive-step rate (train/dev/test): 0.86/0.87/0.86, defined as the rates computed over succeeded calls that are marked decisive (high-confidence, answer-contributing) under our thresholds.
Mean clip duration: 11.8\,s $\pm$ 6.4.
Per-tool usage (train/dev/test, \%):
SEG 24.1/23.8/24.3, ZOOM 17.1/17.8/18.0,
TRK 22.9/22.3/22.1, OCR 20.2/20.0/19.9,
TEMP 10.0/9.7/9.9, PROP 5.9/6.2/5.8.

\noindent\textbf{De-duplication and leakage control.}
Near-duplicates are filtered at image and clip levels via two-stage checks (visual embeddings + perceptual hashes); video checks run on uniformly sampled keyframes (1\,fps). For Pixel TTRL, candidate neighbors flagged by the de-dup filters are masked from the retrieval index, and evaluation media are excluded at ingest; an audit against evaluation whitelists reports 0.00\% overlap (95\% CI [0.00, 0.02]). We release metadata, validators, and replay logs for reproducibility.

\noindent\textbf{Error analysis.}
On a 1{,}500-sample stratified audit, each tool invocation is treated as a binary pass/fail event under the same thresholds as the runtime study. 95\% CIs. Failure rates are: SEG 11.4\% (IoU$<0.5$), TRK 16.9\% (HOTA$<0.15$), OCR 7.5\% (ANLS$<0.85$ after NFKC + punctuation stripping), TEMP 17.3\% (tIoU$<0.5$ or offset $>0.5$\,s), and PROP 9.7\% (attribute mismatch). Around 3\% of audited calls show annotator disagreement; even if all such cases are counted as failures, each rate shifts by less than 1\,pp.

\vspace{-4pt}
\subsection{Training Pipeline}
\label{sec:training}

Pixelis trains a vision–language agent to reason, act, and adapt in pixel space via three phases: SFT learns tool use from Chain-of-Thought–Action traces, CC-RFT steers exploration with curiosity{+}coherence under a KL anchor, and Pixel TTRL performs safe test-time adaptation by answer to behavior voting with KL/EMA stabilization. All phases share a compact tool set (segment object at, get properties, zoom in/out, select frame, read text, track object) with parameterized calls and replayable traces. This action interface complements universal embodied spaces~\cite{zheng2025universalactionsenhancedembodied}, improving pixel-level alignment and reproducibility. Callable tools: SEG, ZOOM, TRK, OCR, TEMP, PROP; each outputs typed arguments and replayable artifacts. Frozen evaluators (for scoring only) include HOTA/CLEAR-MOT (TrackEval), ANLS, and DINOv2~\cite{oquab2023dinov2,luiten2020IJCV}, trying DINOv3~\cite{siméoni2025dinov3}.

\begin{figure*}[t]
    \centering
    \includegraphics[width=0.95\textwidth]{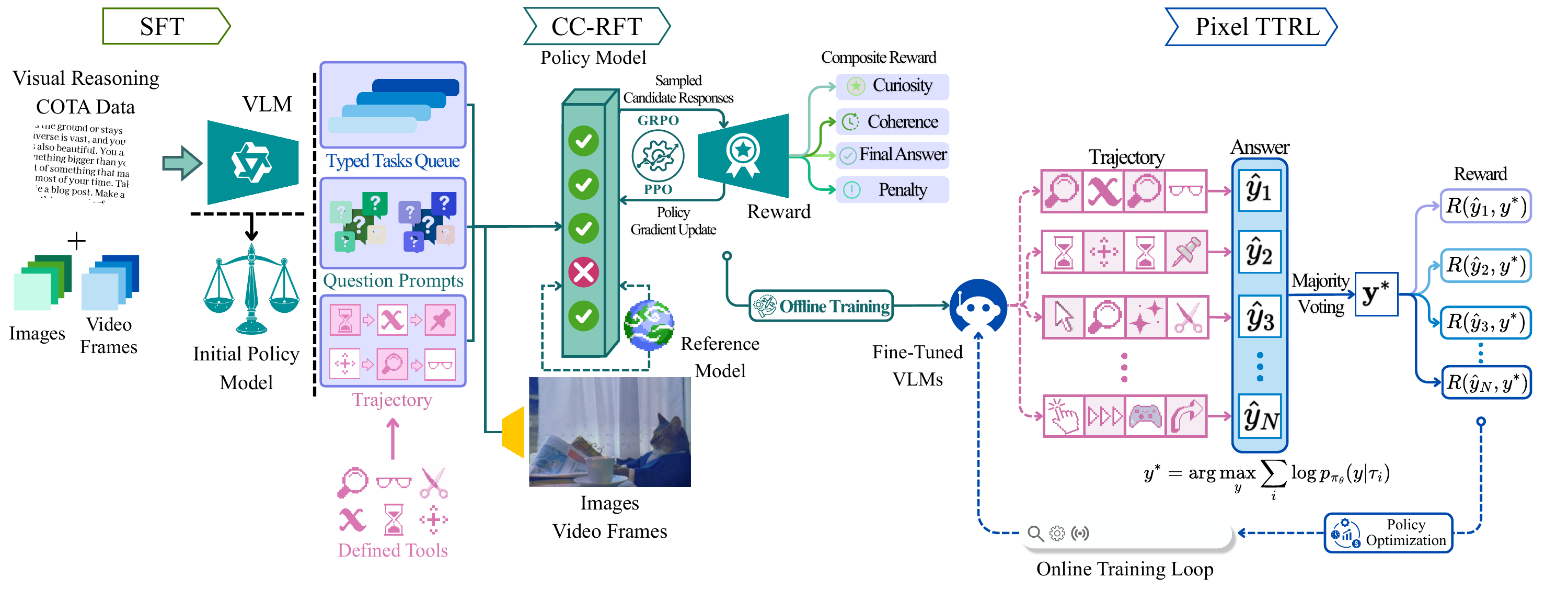}
    \vspace{-6pt}
    \caption{Three-phase training of Pixelis. SFT learns a tool-use grammar; CC-RFT shapes exploration with curiosity, coherence, and a light efficiency prior; Pixel TTRL adapts at test time by retrieving neighbours and updating toward behaviourally consistent trajectories under a KL-to-EMA safety constraint, turning raw tool traces into shorter, structured pixel toolchains.}
    \label{fig:train_pipeline}
\vspace{-4pt}
\end{figure*}

\subsubsection{Phase 1: Supervised Fine-Tuning (SFT)}
\label{subsec:sft}

\noindent\textbf{Goal/data interface.} Each example has an image/video, a prompt, and a CoTA trace with intermixed thought tokens and tool calls (op name + typed arguments). Tool outputs (boxes, masks, tracklets, OCR spans, temporal marks) are serialized back as short, structured tokens (normalized coordinates at $10^{-2}$; subword OCR tiles) so the model conditions on its own actions.

\noindent\textbf{Masked imitation loss.}
SFT uses a standard next-token cross-entropy loss, but we upweight tokens that belong to tool names and arguments (weight $w_{\text{act}}>1$, default $2$) so that mistakes on actions matter more than mistakes on plain text. We still train with teacher forcing, and apply $5\%$ feedback dropout that randomly hides previous tool outputs so the model learns to handle missing or noisy history.

\noindent\textbf{Tool-argument auxiliary loss (on decisive steps $\mathcal{D}$).}
\begingroup
\setlength{\abovedisplayskip}{4pt}
\setlength{\belowdisplayskip}{4pt}
\setlength{\jot}{0pt}
\begin{equation}
\label{eq:sft_tool_aux_compact}
\begin{split}
\mathcal{L}_{\mathrm{tool}}
&=\tfrac{1}{|\mathcal{D}|}\!
\sum_{t\in\mathcal{D}}\!\Big[
\lambda_{\mathrm{box}}\mathrm{SmoothL1}(\hat{b}_t,b_t)
+\lambda_{\mathrm{text}}\mathrm{CE}(\hat{y}_t,y_t)\\[-3pt]
&\quad
+\lambda_{\mathrm{mask}}\!\big(1-\mathrm{Dice}(\hat{M}_t,M_t)\big)
+\lambda_{\mathrm{temp}}\mathrm{BCE}(\hat{u}_t,u_t)
\Big]
\end{split}
\end{equation}
\endgroup

with lightweight heads (single-layer MLP, hidden 256) active only when the tool applies, receiving pooled visual features and tool confidences. The architectural details of the dynamics and auxiliary heads are in Appendix~S2. Total SFT loss:
\vspace{-4pt}
\begin{equation}
\label{eq:sft_total_obj_compact}
\mathcal{L}_{\text{SFT,tot}}
=\mathcal{L}_{\text{SFT}}+\alpha_{\text{tool}}\mathcal{L}_{\text{tool}},
\quad \alpha_{\text{tool}}{=}0.2\ (\ll 1).
\end{equation}
\noindent\textbf{Curriculum.} One warm-up epoch on the full pool, then medium{:}hard sampling $2{:}1$ with hardness binned by tertiles of SFT-predicted error; we advance when EMA-smoothed dev loss $L_k$ satisfies $L_{k+1}\!\le\!1.05\,L_k$. Regularization: label smoothing 0.05, grad-norm cap 1.0, and 2\% dropout on early actions ($t{\le}3$) during teacher forcing. SFT exports $\pi_{\mathrm{SFT}}$ plus token normalizers (running mean/variance over log-probs and entropies) and fixed per-tool scalers that reweight curiosity and penalty terms in CC\mbox{-}RFT/TTRL.

\subsubsection{Phase 2: Curiosity--Coherence RFT (CC\mbox{-}RFT)}
\label{subsec:ccrft}

\noindent\textbf{Objective.}
For a rollout $\tau=(s_0,a_0,\ldots,s_T)$ we maximize
\vspace{-4pt}
\begin{equation}
\label{eq:ccrft_reward}
R(\tau)=w_1R_{\mathrm{final}}+w_2R_{\mathrm{cur}}+w_3R_{\mathrm{coh}}-w_4R_{\mathrm{pen}},
\end{equation}
where $R_{\mathrm{final}}\!\in\!\{+1,-1\}$ and $R_{\mathrm{pen}}$ penalizes invalid actions and chains longer than $L_0$. Curiosity rewards states where the tool-conditioned dynamics head predicts a different next visual state than the current policy, but weights each step by an uncertainty gate so that large prediction errors only count when epistemic variance is low. Coherence sums standardized adjacent cosines of step embeddings to discourage tool hopping:
$R_{\mathrm{coh}}(\tau)=\sum_{t=1}^{T}\mathrm{zscore}\!\big(\cos(E_t,E_{t-1})\big)$. We tune $w_i$ on the dev split; intrinsic terms are batch-wise z-scored to balance magnitudes.

\noindent\textbf{Step embeddings $E_t$.}
Visual tokens $v_t$ are masked-pooled at layer $L_{\text{mm}}$; $x_t$ mean-pools recent thoughts and serialized tool outputs (with confidences when available); $\mathrm{onehot}(a_{t-1})$ encodes the previous action. We project to unit norm
\vspace{-3pt}
\begin{equation}
\label{eq:embed_Et}
E_t=\frac{g_\phi\!\left([\,v_t\ \|\ x_t\ \|\ \mathrm{onehot}(a_{t-1})\,]\right)}
        {\left\|g_\phi\!\left([\,v_t\ \|\ x_t\ \|\ \mathrm{onehot}(a_{t-1})\,]\right)\right\|_2},
\quad \dim(E_t)=512.
\end{equation}
and stop gradients into backbone tokens during CC\mbox{-}RFT/TTRL (only $g_\phi$ updates). Adjacent cosines are z-scored on the dev split per task family and reused at test time. RaCPR uses $E_t$ only for training diagnostics; evaluation uses a frozen external encoder $E^{\text{ext}}$ (DINOv2) on footprint crops.

\noindent\textbf{Policy update.}
We use a GRPO-style policy gradient~\cite{deepseek-math} with a KL penalty to the SFT policy. For each prompt, we keep a small group of top-$K$ trajectories ($K{=}8$), compute group-relative scores, and increase the probability of higher-scoring trajectories. A lightweight PID controller adjusts the KL weight so that the average token-level KL stays in a narrow corridor around $0.15$, keeping updates effective but bounded. Temporal localization follows VTG~\cite{wu2025surveyvideotemporalgrounding}.

\noindent\textbf{Outcome.}
CC\mbox{-}RFT produces shorter, structured, executable toolchains with higher tool fidelity and smoother transitions, without process critics or plan-level supervision; intrinsic summaries are cached as priors and retrieval keys for TTRL.

\subsubsection{Phase 3: Pixel Test-Time RL (Pixel TTRL)}
\label{subsec:ttrl}

\noindent\textbf{Hybrid retrieval.}
For each query $q$, we retrieve a neighborhood $\mathcal{N}(q)$ using a mixture of text and pixel signals: text keys encode the prompt and interim thoughts, and pixel keys encode tool outputs (SEG masks/crops, TRK tracklets, OCR tiles, TEMP segments). The final similarity is a convex combination of text-key and pixel-key similarities, with equal weights by default.

\noindent\textbf{Successful set and exemplar.}
For each query, we sample $N$ rollouts, take the majority answer $\hat a$ under a low-entropy majority ($<0.2$), and form a successful set $\mathcal{S}$ of trajectories that predict $\hat a$. Among these, we choose a single exemplar trajectory that we will update toward: we prefer chains that are shorter, have higher curiosity+coherence summaries, and higher pixel fidelity VisFid$(\tau)$, using a fixed scoring combination shared across all experiments.

\noindent\textbf{Consensus and abstention.}
Instead of hard majority, we use uncertainty-weighted voting
$w_j \propto e^{-H_j}\cdot Cal_j\cdot VisFid(\tau^{(j)})$,
where $H_j$ is token entropy on decisive steps and $Cal_j$ is temperature calibration.
Dawid–Skene (EM) treats rollouts as annotators to estimate reliability; the decision combines answer votes and behavioral alignment. If selective risk is high (conformal set $\mathcal{C}$ from entropy/disagreement with $|\mathcal{C}|{>}1$ or margin $<\delta$), we abstain; KL safety remains active.

\noindent\textbf{Pixel fidelity and pseudo-references.}
For each trajectory, we only score decisive tool steps (SEG/ZOOM, TRK, OCR, TEMP, PROP). At each such step we compare the tool output with a pseudo-reference drawn from other successful trajectories or retrieved neighbours, using the standard metric for that tool (IoU for boxes/masks, average IoU over tracklets, ANLS for OCR, and F1-style scores for temporal and property tools). \texttt{VisFid} is the average of these per-step scores, with pseudo-references chosen by simple matching rules that follow the corresponding public evaluation protocols.

\noindent\textbf{Behavioral similarity.}
Each trajectory $\tau$ is a sequence of tool calls and their pixel footprints. To compare $\tau$ with the exemplar $\tau^\star$, we measure how similar their tool sequences are (edit distance on action names) and how similar their footprints are by softly aligning steps and averaging IoU/ANLS over matched regions. We then take a weighted combination of these two terms; the exact alignment and weights are given in Appx.~S6.1.

\noindent\textbf{Objective.}
Neighborhood value is an EMA over curiosity/coherence; values and pseudo-references are stop-gradient:
\vspace{-6pt}
\begin{equation}
\label{eq:value_pen}
\begin{gathered}
\overline{v}(\mathcal{N}(q))
  = \tfrac{1}{|\mathcal{N}(q)|}\!\sum_{h\in\mathcal{N}(q)} v(h),\\[-2pt]
\mathtt{Pen}(\tau)
  = \alpha_{\text{inv}}\!\sum_t\!\mathbb{I}[a_t\!\notin\!\mathcal{A}_{\text{valid}}]
   + \alpha_{\text{len}}\!\max(0,|\tau|-L_0).
\end{gathered}
\end{equation}
Let $m(q)\!\in\!\{0,1\}$ be an abstention mask ($0$ if low-confidence consensus). With
\vspace{-3pt}
\begin{equation}
r(\tau;\tau^\star)=\mathbb{1}\{\text{ans}(\tau)=\hat a\}
+\kappa\,\texttt{Sim}_{\text{behav}}(\tau,\tau^\star)-\lambda_{\text{pen}}\mathtt{Pen}(\tau),
\end{equation}
baseline $b(q)$ (EMA over neighborhood value), and advantage $A(\tau,q)=r(\tau;\tau^\star)-b(q)$,
the loss is
{\small
\begin{equation}
\label{eq:ttrl_obj}
\mathcal{L}
=-\,\overline{v}(\mathcal{N}(q))\,
\Big[r(\tau;\tau^\star)\Big]\log\pi_\theta(\tau)
+\beta\,\mathrm{KL}_{\text{tok}}\!\big(\pi_\theta\;\|\;\pi_{\text{EMA}}\big),
\end{equation}
}
with $\pi_{\text{EMA}}\!\leftarrow\!\rho\,\pi_{\text{EMA}}+(1{-}\rho)\pi_\theta$ and $\beta$ keeping token-KL within [0.10, 0.20] (target 0.15). When $m(q){=}0$, gradients are masked and only EMA/KL apply, so Pixel TTRL provides test-time supervision without a process critic. Defaults and ranges for $\gamma$, top-$K$, de-dup thresholds, and gates are in Appx.~S6.1.

\subsection{Design Choices and Interfaces}
\label{subsec:design}
Curiosity gating suppresses spurious novelty on stochastic textures, and local coherence curbs tool hopping without plan-level critics; answer–behavior voting aligns answers with box/mask/track/OCR evidence and selects a short, high-fidelity exemplar to keep chains concise and auditable. Interfaces: SFT to RFT via a KL anchor plus token statistics; RFT to TTRL via intrinsic summaries as priors and retrieval keys; TTRL to continual competence via value-aware updates under a KL–EMA corridor.
\captionsetup{skip=4pt}
\begin{table*}[t]
\centering
\footnotesize
\setlength{\tabcolsep}{2pt}
\setlength{\aboverulesep}{1pt}
\setlength{\belowrulesep}{1pt}
\renewcommand{\arraystretch}{0.94}
\caption{Results on public benchmarks. We compare Pixelis to answer-only self-consistency, VLM test-time adaptation methods, process-supervised baselines, and a replicated Pixel Reasoner variant for each training phase. The upper block lists proprietary, non-directly comparable models; all claims use the same 8B backbone. Scores are reported on Tool-needed subsets with matched compute, sampling budget, acceptance rate, and token-level KL; {\small $\dagger$} denotes a win on the Tool-needed subset.}
\label{tab:comprehensive_comparison_expanded}
\begin{tabularx}{0.95\textwidth}{>{\raggedright}X l c c c c c c}
\toprule
\makecell[l]{Model \\ Metric} & \textbf{Size} & \makecell[c]{\textbf{V*} \\ \textbf{Bench~\cite{vstar}}} & \makecell[c]{\textbf{MMBench} \\ \textbf{v1.1 en~\cite{liu2024mmbench}}} & \makecell[c]{\textbf{MVBench~\cite{li2024mvbench}}} & \makecell[c]{\textbf{InfoVQA} \\ \textbf{test~\cite{DBLP:journals/corr/abs-2104-12756}}} & \makecell[c]{\textbf{Video-} \\ \textbf{MMMU~\cite{hu2025videommmu}}} & \makecell[c]{\textbf{VSI} \\ \textbf{Bench~\cite{yang2024think}}} \\
\midrule
GPT-5 (minimal)           & -    & -    & 81.4 & 69.9 & -    & 61.6 & -    \\
Gemini-2.5-Pro (minimal)  & -    & -    & 86.6 & 82.9 & -    & 79.4 & 31.4 \\
Seed-1.5-VL~\cite{guo2025seed15vl}               & 20B  & 89.5 & 88.0 & 74.3 & 89.3 & 72.1 & -    \\
Qwen3-VL-8B-Instruct~\cite{qwen3vl2025launch}      & 8B   & 86.4 & 85.0 & 68.7 & 83.1 & 65.3 & 59.4 \\
Qwen3-VL-30B-A3B-Instruct~\cite{qwen3vl2025launch}  & 30B  & 89.5(w tools) & 87.0 & 72.3 & 82.0 & 68.7 & 63.2 \\
InternVL3.5-A3B (with tools)~\cite{wang2025internvl35advancingopensourcemultimodal} & 30B  & - & 84.8 & 72.1 & - & - & 63.7 \\
\midrule
\multicolumn{8}{c}{Ours vs Baseline} \\
\midrule
Pixelis (Qwen3-VL-8B-Instruct) & 8B & 90.1 & 89.5 & 73.8$^{\dagger}$ & 87.9$^{\dagger}$ & 69.8$^{\dagger}$ & 64.4$^{\dagger}$ \\
Pixel Reasoner (Qwen3-VL-8B-Instruct) & 8B & 88.3 & 86.9 & 69.8 & 85.8 & 66.5 & 60.1 \\
\midrule
\multicolumn{8}{c}{Additional baselines — answer-level (no tools/updates)} \\
\midrule
RV Self-Consistency (answer-only)~\cite{wang-etal-2025-ranked} & 8B & 86.9 & 86.0 & 68.4 & 83.6 & 65.1 & - \\
\midrule
\multicolumn{8}{c}{Additional baselines — VLM test-time adaptation (no tools, with adaptation)} \\
\midrule
Realistic TTA of VLMs (StatA)~\cite{zanella2025stata} & 8B
& - & 86.1 & 69.5 & 82.7 & 66.2 & - \\
RA-TTA (retrieval-augmented)~\cite{ra-tta-iclr25} & 8B
& - & 86.4 & 70.6 & 84.0 & 66.8 & - \\
\midrule
\multicolumn{8}{c}{Process-supervised (tools) in main text} \\
\midrule
PRM (process reward; tools; 8B)~\cite{zhao2025genprm, lightman2023letsverify} & 8B & 88.9 & 87.8 & 71.1 & 86.2 & 67.6 & 62.3 \\
Step Self-Consistency (step-level)~\cite{wang2022selfconsistency} & 8B & 88.0 & 87.2 & 70.5 & 85.4 & 66.9 & - \\
Late-fusion (step logits, equal-variance) & 8B & 87.1 & - & - & 85.6 & 67.0 & - \\
\midrule
\multicolumn{8}{c}{Ablation} \\
\midrule
Pixelis: SFT only / RFT only              & 8B & 87.7/86.9 & 87.6/86.2 & 68.6/70.7 & 84.9/84.4 & 67.4/66.5 & 60.1/60.8 \\
Pixelis: TTRL only             & 8B & 86.8 & 85.9 & 69.6 & 83.5 & 66.1 & 59.6 \\
\makecell[l]{Pixelis: SFT + RFT /\\ SFT + TTRL} & 8B & 89.2/88.5 & 88.6/88.3 & 71.8/70.9 & 86.7/86.1 & 68.5 & 62.8 \\
Pixelis: RFT + TTRL            & 8B & 88.1 & 87.9 & 70.9 & 85.8 & 67.8 & 61.9 \\
\bottomrule
\end{tabularx}
\vspace{-8pt}
\end{table*}

\section{Experiments}
\label{sec:experiments}

We evaluate Pixelis on six public benchmarks stressing spatial/temporal reasoning. Backbone: Qwen3-VL-8B-Instruct~\cite{qwen3vl2025launch}. We organize by three questions; unlike long-video agents with semantic actions~\cite{liu2025videomindchainofloraagentlong}, we emphasize executable pixel chains and process audits.

\vspace{2pt}
\subsection{Evaluation setup and metrics}
\label{subsec:eval_setup}
\noindent\textbf{Data protocol.}
Test images/clips are SHA\mbox{-}256 de\mbox{-}duplicated, and a curated auditing split includes pixel references (boxes, masks, OCR spans, tracklets) with replayable traces. Retrieval\mbox{-}key near\mbox{-}duplicates are suppressed by a pHash prefilter plus CLIP\mbox{-}embedding NMS; same\mbox{-}source or highly similar keys are masked. Cross\mbox{-}split near\mbox{-}duplicate candidates before filtering are 0.41\% (images) / 0.36\% (clips); the two\mbox{-}stage filter removes 0.37\% / 0.33\%, leaving a residual $\approx$0.05\% (95\% CI [0.03, 0.08], bootstrap over held\mbox{-}out shards). For evaluation leakage, comparing test media to the TTRL index yields 0/50{,}216 exact overlaps and 12/50{,}216 near\mbox{-}duplicate candidates, all masked at ingest; effective overlap is 0.00\% (95\% CI [0, 0.007\%]).

\noindent\textbf{Task and tool metrics.}
We report Accuracy and ANLS (InfoVQA) plus pixel/tool fidelity when applicable: IoU (boxes/masks), Boundary\mbox{-}F1, CER for OCR, HOTA/CLEAR\mbox{-}MOT for tracking; temporal localization follows VTG survey guidance~\cite{wu2025surveyvideotemporalgrounding}.

\noindent\textbf{Process metrics (defined here).} Beyond task scores, we quantify how answers are obtained. For Rate of Pixel Reasoning (RaPR), let $u_t\!\in\!\{0,1\}$ indicate a valid visual–tool step at time $t$. For trajectory $h$ of length $T$, \vspace{-8pt} \begin{equation} \mathrm{RaPR}(h)=\tfrac{1}{T}\sum_{t=1}^{T} u_t \,,\quad \overline{\mathrm{RaPR}}=\mathbb{E}_{h}[\mathrm{RaPR}(h)]. \end{equation} Validity uses external verifiers (IoU/Boundary-F1, ANLS, HOTA/CLEAR-MOT; VTG overlaps). \footnote{TrackEval implements HOTA/CLEAR-MOT and related evaluators; we use it as a frozen scorer, not as a tracker.} For Rate of Composite Pixel Reasoning (RaCPR), we detect coherent multi-step chains by gating adjacency of step embeddings with fixed $z$-scoring. Let $c_t=\cos\!\big(E_{\text{ext}}(\pi_t),E_{\text{ext}}(\pi_{t-1})\big)$, $\tilde c_t=z(c_t)$ (dev-split, per-task mean/STD), and $g_t=\mathbf{1}\{u_{t-1}{=}1,u_t{=}1,\tilde c_t\!\ge\!\tau\}$. Consecutive $g_t{=}1$ form candidates $\{C_k\}$; keep $|C_k|\!\ge\!L_{\min}$. For a qualified chain $C$, \vspace{-4pt} \begin{equation} \small q(C)=\frac{1}{|C|}\sum_{t\in C}\!\big[\tilde c_t-\tau\big]_+ -\alpha_{\text{len}}\frac{(|C|-L_0)_+}{|C|} -\alpha_{\text{inv}}\frac{\sum_{t\in C}(1-u_t)}{|C|}, \end{equation} and we report $\mathrm{RaCPR}(h)=\max_{C} q(C)$ and $\overline{\mathrm{RaCPR}}=\mathbb{E}_h[\mathrm{RaCPR}(h)]$. For evaluation, cosines are computed using a frozen external encoder $E_{\text{ext}}$ (DINOv2), decoupling RaCPR from the trained step embeddings $E_t$. Uncertainty is reported as BCa 95\% confidence intervals with Benjamini–Hochberg correction across tasks and hyperparameter settings $(\tau, L_{\min}, \gamma)$.

\noindent\textbf{Answer-level strong baseline.} We add a compute-matched self-consistency baseline with temperature-scaled confidence and abstention (“Ans-SSC (cal+abstain)”) in Table~\ref{tab:comprehensive_comparison_expanded}.

\noindent\textbf{Sensitivity and human correlation.}
With dev-split $z$-scoring on a frozen external encoder ($E_{\text{ext}}{=}$DINOv2; we obtained similar trends with CLIP-L), $\overline{\mathrm{RaCPR}}$ varies by at most $\pm0.5$ over $\tau\!\in[0.25,0.40]$, $L_{\min}\!\in\!\{2,3,4\}$, $\gamma\!\in[0.15,0.30]$, and task Accuracy varies by at most $\pm0.2$ (defaults in Tab.~\ref{tab:hp-rcpr}). On 200 trajectories (2 annotators, 5-pt scale), Spearman $\rho$ (BCa 95\% CIs) is: RaPR $0.62$ [$0.55$, $0.68$], RaCPR $0.57$ [$0.49$, $0.64$] (BH-corrected).

\begin{table}[t]
\centering
\scriptsize
\setlength{\tabcolsep}{2pt}
\renewcommand{\arraystretch}{0.92}
\caption{De-duplication and leakage audit. We report the number of near-duplicate image/clip pairs across CoTA splits and between each split and the TTRL retrieval index; low counts indicate limited train–eval leakage.}
\label{tab:dedup_audit}
\vspace{-3pt}
\begin{adjustbox}{max width=\columnwidth}
\begin{tabular}{@{}lccc@{}}
\toprule
Split & Candidates (\%) & Removed (\%) & Residual, 95\% bootstrap CIs (\%) \\
\midrule
Train vs Dev  & 0.41 / 0.37 & 0.36 / 0.31 & 0.05 [0.03, 0.08] / 0.05 [0.03, 0.08] \\
Train vs Test & 0.38 / 0.34 & 0.36 / 0.29 & 0.04 [0.02, 0.07] / 0.04 [0.02, 0.07] \\
\midrule
\multicolumn{4}{@{}p{\columnwidth}@{}}{%
\textit{Eval vs TTRL index:}
0 / 50{,}216 exact;\;12 / 50{,}216 candidates;\;
0.00\% [0, 0.007\%].
}\\
\bottomrule
\end{tabular}
\end{adjustbox}
\vspace{-12pt}
\end{table}

\noindent\textbf{Fairness and compute.}
We re-run open models with official scripts; APIs use public protocols (rate/budget capped), reporting the best of three seeds when applicable.
SFT: 3 epochs, 2:1 medium:hard.
RFT: GRPO (Group-Relative Policy Optimization) ($K{=}8$), target token-KL $\approx 0.15$.
TTRL: neighborhood $k{=}8$, $N{=}8$ rollouts, EMA $\rho{=}0.99$, grad-norm cap 1.0; arbitration/de-dup/decisive-step/soft-DTW follow Sec.~\ref{subsec:ttrl} Inline settings. Online runs for 8K updates.”

\begin{figure}[ht!]
\centering
\includegraphics[width=\linewidth]{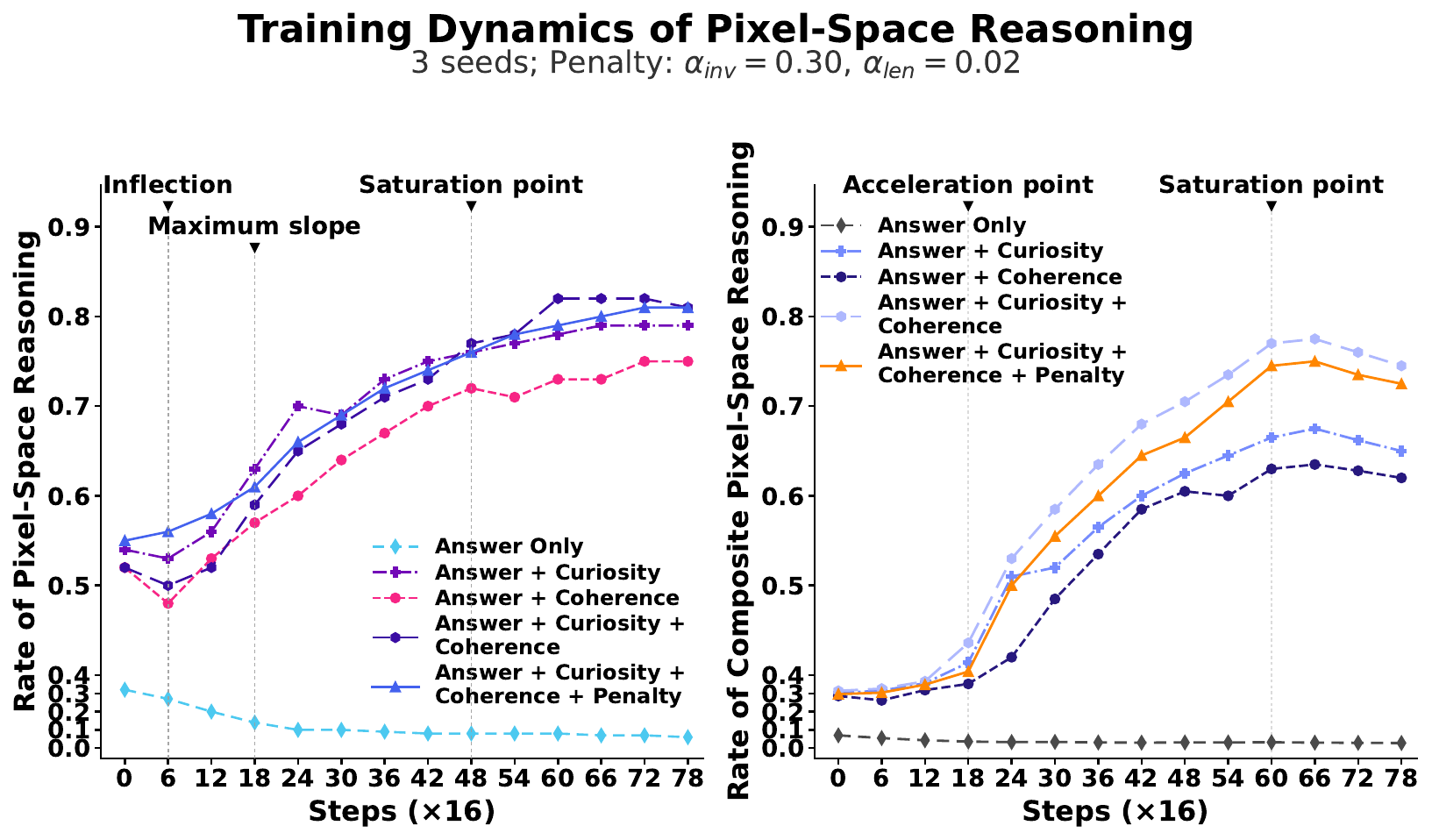}
\caption{RFT process metrics: RaPR (top) and RaCPR (bottom). We compare Answer Only, +Curiosity, +Coherence, +Curiosity+Coherence, and +Curiosity+Coherence+Penalty (Pixelis). Adding curiosity alone increases RaPR but hurts RaCPR; adding coherence and a light penalty yields the highest RaPR/RaCPR with lower variance across seeds. Bars show means over 3 seeds with 95\% BCa bootstrap confidence intervals (BH-corrected).}
\label{fig:rft_dual}
\end{figure}

\subsection{RQ1: Does CC-RFT yield structured and exploratory reasoning?}
\label{subsec:rq1}
Under same-backbone controls, matched acceptance, and a KL-to-EMA corridor, ablating reward terms within the same GRPO setup shows: removing coherence (RFT-Curiosity) degrades chain stability (RCS, run-to-run chain stability) and tracking fidelity (MOTA) with visible tool hopping; removing curiosity (RFT-Coherence) reduces exploration (AE, action entropy) and lowers accuracy via redundant, over-local chains; the full CC-RFT improves Accuracy/ANLS and lifts RaPR/RaCPR with smaller variance, indicating concise, coherent toolchains rather than trial-and-error. Figure~3 reports RaPR/RaCPR dynamics; Table~\ref{tab:comprehensive_comparison_expanded} summarizes end-point scores. Replacing adjacency-based coherence with a generic smoothness prior (L2 on consecutive arguments, stop-gradient) recovers only 31\% of the RaCPR gain and underperforms Accuracy by 0.7 at matched KL, suggesting adjacency on $E_t$ provides structure beyond generic regularization. If jitter appears, raise the adjacency gate $\tau$ before increasing penalties; if exploration stalls, slightly lower $\tau$ while keeping $L_{\min}$ (defaults in the appendix).

\begin{figure}[ht!]
\centering
\includegraphics[width=\linewidth]{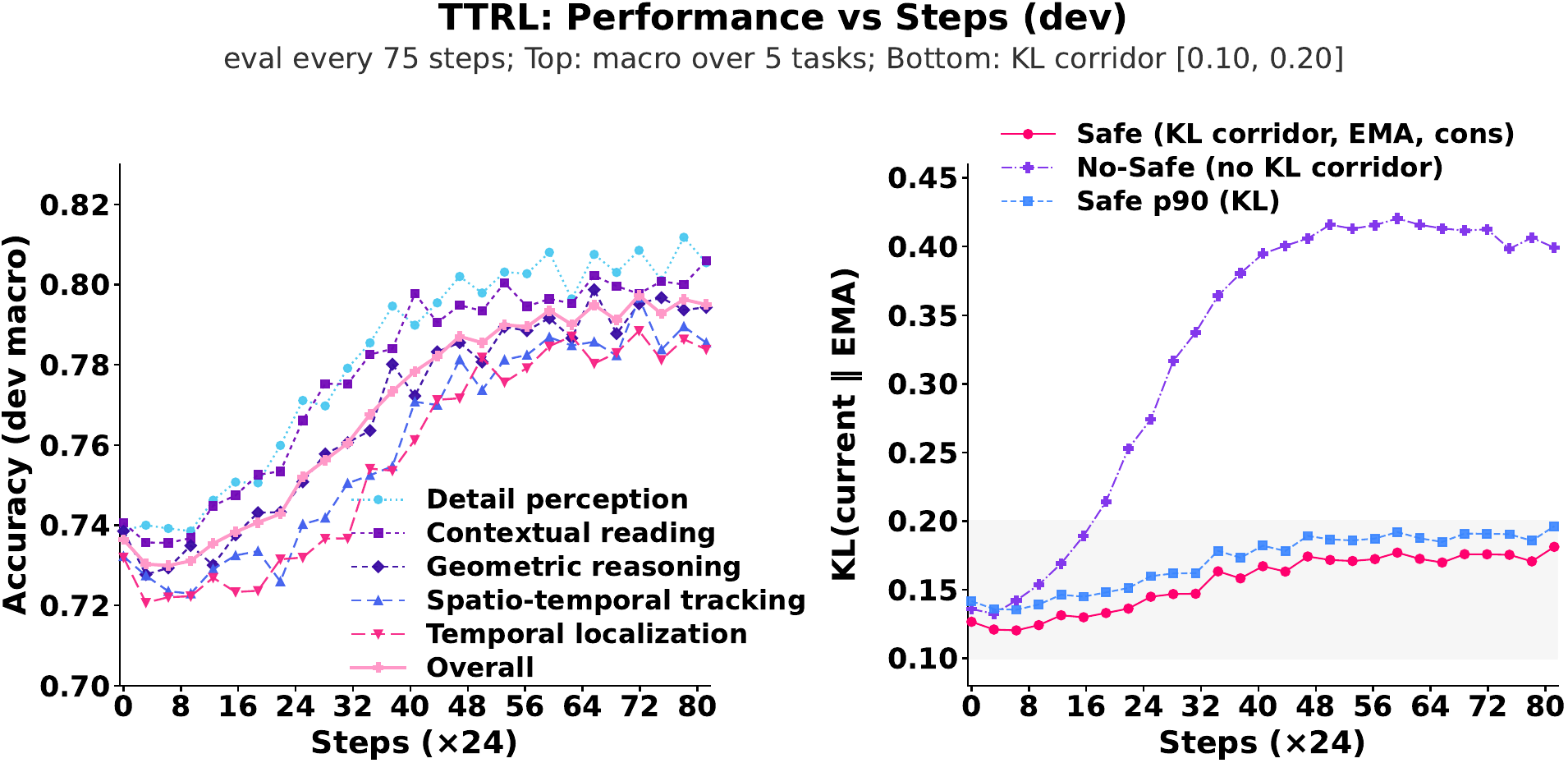}
\caption{Pixel TTRL: accuracy (top) and token-KL to EMA (bottom) within corridor $[0.10,0.20]$. Safe variant (value-aware retrieval, trajectory voting, EMA+KL) stays in-corridor; no-safety drifts and degrades. PID-controlled $\beta$ keeps KL bounded.}
\label{fig:ttrl_dual}
\vspace{-8pt}
\end{figure}

\vspace{-3pt}
\subsection{RQ2: Can Pixel TTRL adapt safely under unlabeled shift?}
\label{subsec:rq2}
Under lighting and motion shifts, with same-backbone controls, matched acceptance, and a KL-to-EMA corridor, Pixelis-Online improves accuracy from 73.0 to 76.5 within 8k updates ($p{<}0.01$, two-sided permutation, BH across 6 tasks; protocol in Appx.~S8) while token-KL stays near the corridor target (median $\approx$0.16, P95 $\approx$0.19). Value-aware reweighting yields about $1.6\times$ more accepted updates per accuracy point. Removing KL control drives token-KL above 0.40 with accuracy collapse; dropping the EMA anchor inflates gradient variance. RaCPR increases as trajectory voting prefers short, well-structured exemplars. Under adversarial-majority stress (flipped answers or semantically similar but label-mismatched neighbors), hard-majority TTRL drifts, with accuracy decreasing and RaCPR variance increasing, whereas uncertainty-weighted consensus with abstention keeps KL well below the no-safety variant and improves accuracy. Sweeping the abstention margin $\delta$ produces a risk–coverage curve where Err@Sel drops from $8.7\%$ to $7.1\%$ at roughly $9$–$12\%$ lower coverage (Tab.~\ref{tab:risk_coverage}). For plan re-execution robustness (PRC) under JPEG compression $q{=}10$–$30$, $\pm 2\%$ resize, and $\pm 1$ frame jitter, accuracy improves from $78.9\%$ to $83.4\%$, with RaCPR correlating with PRC ($\rho{=}0.51$ [0.44, 0.58]).

\begin{table}[ht!]
\centering
\scriptsize
\setlength{\tabcolsep}{2pt}
\renewcommand{\arraystretch}{0.9}
\caption{RaCPR hyperparameters shared across benchmarks. Sensitivity within $\tau\!\in\![0.25,0.40]$, $L_{\min}\!\in\!\{2,3,4\}$, $\gamma\!\in\![0.15,0.30]$.}
\label{tab:hp-rcpr}
\vspace{-3pt}
\begin{adjustbox}{max width=\columnwidth}
\begin{tabular}{@{}llllll@{}}
\toprule
Symbol & Value & Symbol & Value & Symbol & Value \\
\midrule
$\tau$ (adj.\ gate) & 0.30 &
$L_{\min}$ (min chain) & 3 &
$L_0$ (len prior) & 6 \\
$\gamma$ (soft-DTW temp.) & 0.20 &
$\alpha_{\text{len}}$ (len pen.) & 0.02 &
$\alpha_{\text{inv}}$ (inv.\ pen.) & 0.30 \\
\bottomrule
\end{tabular}
\end{adjustbox}
\vspace{-6pt}
\end{table}

\vspace{-8pt}
\begin{table}[ht!]
\centering
\scriptsize
\setlength{\tabcolsep}{3pt}\renewcommand{\arraystretch}{0.9}
\caption{Selective adaptation on $5{,}000$ shifted queries. Coverage is the answered fraction; Err@Sel is the error on that subset. Weighted voting with abstention lowers Err@Sel and KL P95 relative to hard majority.}
\label{tab:risk_coverage}
\vspace{-2pt}
\begin{tabular}{lccc}
\toprule
Strategy & Coverage (\%) & Err@Sel (\%) & KL P95 \\
\midrule
Hard majority & 100.0 & 8.7 & 0.29 \\
Weighted + abstention ($\delta{=}0.08$) & 88.9 & 7.1 & 0.19 \\
Hard majority (adv.\ 30\% flipped) & 100.0 & 11.3 & 0.31 \\
Weighted + abstention (adv.\ $\delta{=}0.10$) & 86.2 & 9.2 & 0.21 \\
\bottomrule
\end{tabular}
\vspace{-6pt}
\end{table}

\subsection{RQ3: Do executable pixel tools unlock capabilities beyond text abstraction?}
\label{subsec:rq3}
\vspace{-3pt}
Across six benchmarks with same-backbone controls, matched acceptance, and a KL-to-EMA corridor, RFT-Full outperforms RFT-Base and our re-implementation of Pixel Reasoner, and Pixelis-Online adds further gains while staying in-corridor (p$<$0.01, two-sided permutation, BH-corrected). RaPR/RaCPR and decisive-step IoU/ANLS/MOTA improve jointly, and audits show fewer ID switches, more reliable OCR-before-compare, and precise decisive chains (Fig.~6). Pixel-grounded retrieval explains the trend: pixel keys (boxes, masks, tracks, OCR tiles) align similarity with verifiable evidence, defining Visual Fidelity (VisFid) as the fraction of retrieved exemplars whose evidence matches the query footprint. Sweeping the mixing weight $\lambda_{\text{pix}}$ (pixel vs.\ semantic keys, default 0.5) yields coupled gains in VisFid and RaCPR, while $\lambda_{\text{pix}}>0.85$ slightly hurts cross-semantic generalization on V*Bench, suggesting a balanced mix. Answer-only self-consistency improves accuracy but cannot lift RaPR/RaCPR; VLM TTA without tools gains at test time but drifts without executable evidence; process-supervised tool baselines (PRM, step-level self-consistency) help locally, whereas our trajectory-level voting with KL/EMA stabilizes full chains into shorter, coherent plans and safer online updates.

\begin{figure}[t]
\centering
\includegraphics[width=\linewidth]{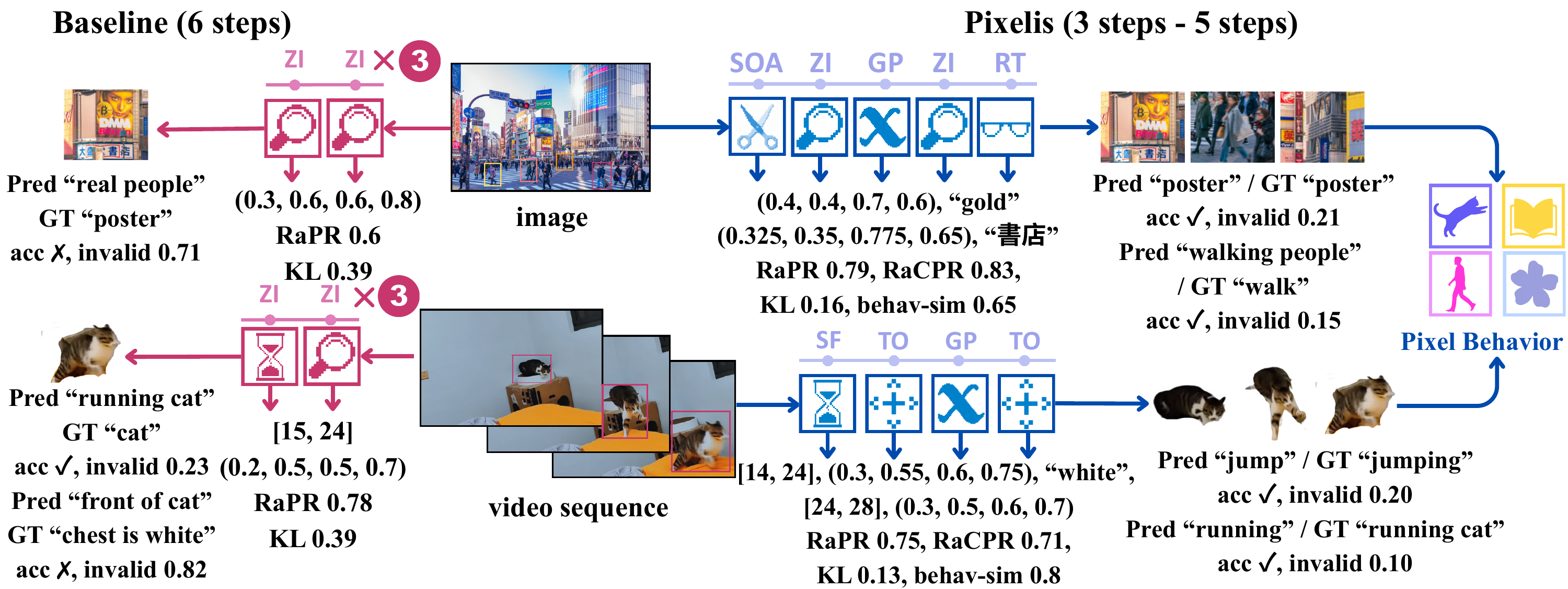}
\caption{Qualitative comparison. The baseline often loops or over-zooms on irrelevant regions, while Pixelis forms shorter, more coherent toolchains that align with the queried evidence, reflected in higher RaPR/RaCPR and VisFid.}
\label{fig:qualitative_examples}
\end{figure}
\vspace{-4pt}

\begin{figure}[t]
  \centering
  \setlength{\abovecaptionskip}{2pt}
  \setlength{\belowcaptionskip}{-6pt}
  \includegraphics[width=\linewidth]{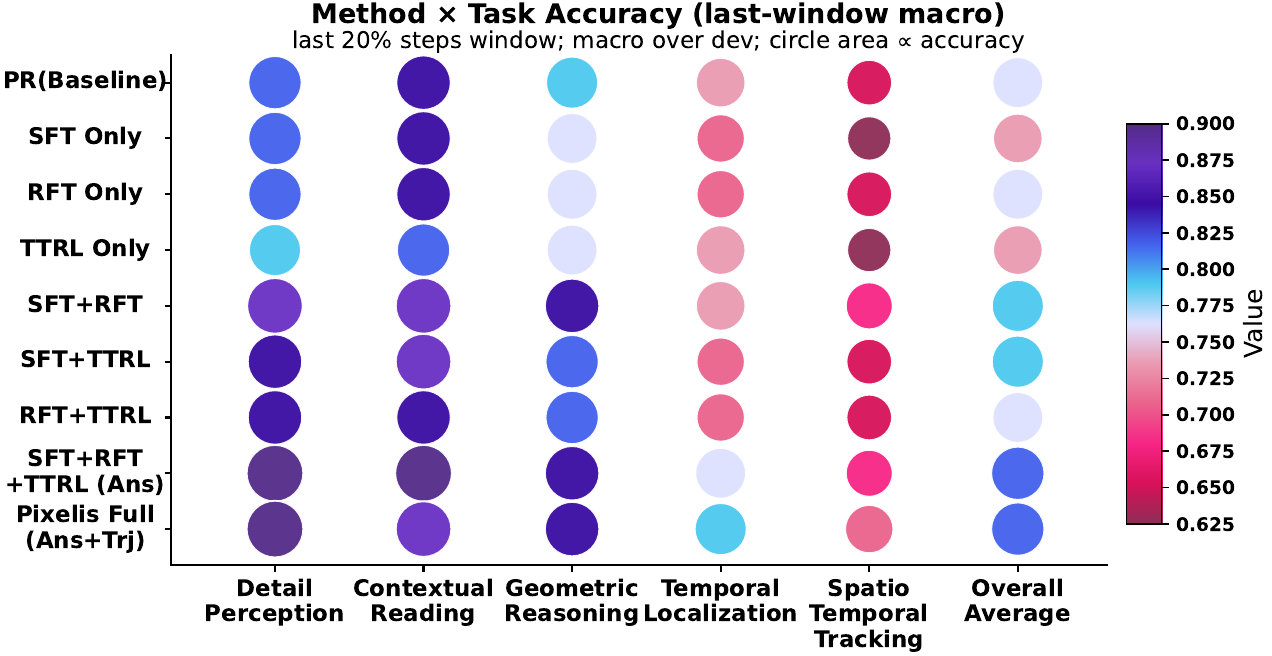}
  \caption{Qualitative behaviors. Coherence regularization prevents oscillatory tool loops and encourages consistent steps, while Pixel TTRL uses retrieved trajectories to self-correct toolchains.}
  \label{fig:qualitative_examples_more}
  \vspace{-6pt}
\end{figure}

\begin{table}[h]
\centering
\scriptsize
\setlength{\tabcolsep}{2pt}
\renewcommand{\arraystretch}{0.95}
\caption{Runtime by tool phase for the 8B model (medians over $1{,}500$ eval queries; latency excludes model forward). “Tool calls/ex (succ.\ rate\%)” is the median number of successful tool invocations per example and their pass rate under fixed thresholds. Attempts also counts failed calls.}
\label{runlatency}
\vspace{-3pt}
\begin{adjustbox}{max width=\columnwidth}
\begin{tabular}{@{}l
                >{\centering\arraybackslash}p{0.17\columnwidth}
                >{\centering\arraybackslash}p{0.17\columnwidth}
                >{\centering\arraybackslash}p{0.25\columnwidth}@{}}
\toprule
Component & Latency (s) & Share (\%) & Tool calls/ex (succ.\ rate\%) \\
\midrule
SEG/ZOOM (mask/bbox)     & 1.17 & 23.1 & 0.9 (88.6) \\
TRK (eval pipeline)      & 1.42 & 28.2 & 0.7 (83.1) \\
OCR (tiles; ANLS)        & 0.39 &  7.7 & 1.0 (92.5) \\
TEMP (VTG)               & 1.57 & 31.0 & 0.4 (82.7) \\
PROP (property)          & 0.16 &  3.2 & 0.7 (90.3) \\
Retrieval+voting (TTRL)  & 0.34 &  6.8 & -- (n/a) \\
\midrule
Total tool-side overhead & 5.05 & 100.0 & -- \\
\addlinespace[2pt]
\multicolumn{4}{@{}p{\columnwidth}@{}}{\textit{End-to-end latency (incl.\ model fwd):} p50 = 5.8\,s,\ p90 = 8.1\,s,\ p95 = 10.2\,s;\ \textit{Avg.\ chain length (decisive steps)} $\bar{L}$ = 3.7}. \\
\multicolumn{4}{@{}p{\columnwidth}@{}}{%
\textit{Chain length by subset:}
$\bar{L}_{\text{Need-Tool}}=\mathbf{4.0}$,\quad
$\bar{L}_{\text{No-Tool}}=\mathbf{2.1}$;\quad
Tool-needed queries cover $\approx$84\% of the eval set.} \\
\bottomrule
\end{tabular}
\end{adjustbox}
\vspace{-4pt}
\parbox{\columnwidth}{\scriptsize
\vspace{2pt}
\textit{Notes.} Full configs and per-tool breakdown with 95\% CIs are in Appx.~S5.2; Retrieval+voting runs once per query and counts toward tool-side and end-to-end latency. Thresholds: SEG IoU$\!\ge$0.5, TRK HOTA$\!\ge$0.15, OCR ANLS$\!\ge$0.85, TEMP tIoU$\!\ge$0.5}
\vspace{-8pt}
\end{table}

\subsection{Ablations and further analyses}
\label{subsec:further}
\noindent Sensitivity scans show $\tau$ trades recall and precision of composite chains, $L_{\min}$ suppresses short bursts, and $\gamma$ tunes alignment; defaults near $\tau{=}0.30$, $L_{\min}{=}3$, $\gamma{=}0.20$ lie at the knee, with Accuracy varying within $\pm 0.2$. Replacing behavioral voting with answer-only self-consistency slows adaptation and inflates RaCPR variance; hard majority is brittle under noisy neighborhoods, whereas uncertainty-weighted voting with abstention prevents catastrophic updates at negligible coverage cost. Under Gaussian noise and occlusion, RFT-Full degrades less and maintains higher RaCPR than Pixel Reasoner; with 30\% vote noise, naive majority TTRL drops $-4.6$ RaCPR versus $-1.2$ for our variant; and cross-dataset transfer (train COCO/MOT-style~\cite{lin2014microsoft,zhang2020simple}, test LVIS-style) preserves a fraction of in-domain performance~\cite{gupta2019lvis}, with RFT-Full outperforming RFT-Base on task and process metrics.

\vspace{-4pt}
\subsection{Reproducibility}
\label{subsec:repro}
We release configs for SFT/CC-RFT/TTRL, checkpoints, RaPR/RaCPR scripts, and logs/replays for tool-call audits and end-to-end online updates. At 8B, total training cost is 805.2 GPU·h. Runtime is dominated by pixel tools ($\approx 87\%$ of latency, mainly segmentation and tracking); online adaptation adds $\approx 0.35$\,s median for retrieval/voting, and KL/EMA bookkeeping is $<0.1$\,s.

\vspace{-8pt}
\section{Conclusion and Limitations}
\label{sec:conclusion}
\vspace{-3pt}
Pixelis learns to act and adapt in pixel space by composing a small set of executable tools. Its three stages play roles: SFT recovers tool syntax and usage patterns from traces; CC-RFT steers exploration with adjacent-step coherence under a mild KL anchor; and Pixel TTRL carries out test-time adaptation through trajectory-level voting with KL–EMA stabilization. Together they yield shorter, more stable, and auditable toolchains while keeping adaptation drift controlled, turning pixel-grounded behavior into supervision for transparent, reproducible agents relevant to embodied AI and human–AI collaboration. Limits remain. The phases are coupled through reused statistics and KL anchoring rather than a single end-to-end objective, so residual SFT bias can persist and Pixel TTRL may under-adapt or oscillate under abrupt shift; process signals can overfit stale behaviors or chase high-entropy textures; coherence is local; and non-differentiable tools (segmentation/OCR/tracking) are brittle on thin structures, stylized fonts, dense layouts, or motion blur, with errors propagating into RaCPR. Practical mitigations include mild end-to-end regularization with adaptive KL budgets, diversity-aware replay with uncertainty gating and rarity upweighting, checks, confidence-based fallbacks, and tool-noise simulation.

{
    \small
    \bibliographystyle{ieeenat_fullname}
    \bibliography{main}
}

\clearpage
\setcounter{page}{1}
\maketitlesupplementary
\setcounter{section}{0}
\setcounter{figure}{0}
\setcounter{table}{0}
\setcounter{equation}{0}
\renewcommand{\thefigure}{S\arabic{figure}}
\renewcommand{\thetable}{S\arabic{table}}
\renewcommand{\theequation}{S\arabic{equation}}

\section{Step embeddings and dynamics head details}
\label{supp:et_dynamics}
\setlength{\parindent}{0pt}
\setlength{\parskip}{2pt}

\textbf{Multimodal layer $L_{\text{mm}}$ and feature extraction.}
We compute visual features at the \emph{3rd block from the top} of the multimodal fusion stack (Qwen3\mbox{-}VL\mbox{-}8B\mbox{-}Instruct), i.e., after cross\mbox{-}attention but before final LN. Denote its visual token grid as $F\in\mathbb{R}^{H'\times W'\times d}$ with stride 16 relative to the input frame. For each decisive step footprint (box/mask/track/temporal slice), we form a binary mask $M\in\{0,1\}^{H'\times W'}$ by rasterizing at the token grid; for temporal spans we pool per\mbox{-}frame then average across frames. Visual summary:
\[
v_t=\frac{1}{\max(1,\sum M)}\sum_{i,j} M_{ij}\,F_{ij}\in\mathbb{R}^{d}.
\]
We do \emph{masked mean pooling} only (no max), with a \(3\times 3\) dilated kernel for thin structures when \(\sum M<9\). Token stride and pooling are fixed; no gradients pass to $F$ during CC\mbox{-}RFT/TTRL.

\textbf{Text/tool features $x_t$.}
We concatenate: (i) the mean of last 32 thought tokens; (ii) serialized tool outputs (normalized coords at $10^{-2}$, OCR tiles hashed to subword ids then averaged); (iii) per\mbox{-}tool confidences when available. A linear projector maps this to \(\mathbb{R}^{d}\).

\textbf{Step embedding $E_t$ (final form).}
Let \(\mathrm{onehot}(a_{t-1})\in\mathbb{R}^{|\mathcal{A}|}\). We use a 2\mbox{-}layer MLP \(g_\phi:\mathbb{R}^{2d+|\mathcal{A}|}\!\to\!\mathbb{R}^{512}\) with GELU and LayerNorm:
\[
E_t=\frac{g_\phi\big([\,v_t \,\|\, x_t \,\|\, \mathrm{onehot}(a_{t-1})\,]\big)}
{\big\|g_\phi\big([\,v_t \,\|\, x_t \,\|\, \mathrm{onehot}(a_{t-1})\,]\big)\big\|_2},
\quad \dim(E_t)=512.
\]
We \emph{stop\mbox{-}grad} into the backbone and text projector; only \(\phi\) is trained in CC\mbox{-}RFT. Cosines \(\cos(E_{t+1},E_t)\) are z\mbox{-}scored per task family using running stats from the \emph{training split} and reused at test time.

\textbf{KL anchor.} During CC-RFT and TTRL, we constrain policy updates 
by $\text{KL}(\pi_\theta \parallel \pi_{\text{SFT}}) \le 0.2$ where 
$\pi_{\text{SFT}}$ is frozen as a reference and reuse the same [0.10, 0.20] corridor as in the main text.

\textbf{Architecture and size of $g_\phi$.}
MLP layers: \([2d{+}|\mathcal{A}|]\!\to\!768\!\to\!512\) with GELU, dropout 0.1, LayerNorm on output. Parameter count \(\approx 1.2\)M (negligible vs backbone). We found deeper MLPs did not help under the KL budget.

\textbf{Tool\mbox{-}conditioned dynamics head (for curiosity).}
Given \([v_t \,\|\, x_t \,\|\, \mathrm{onehot}(a_t)]\), a small MLP predicts the \emph{next} visual summary \(\hat v_{t+1}\) and a validity logit \(\hat r_t\) for the issued tool:
\[
h_\psi:\mathbb{R}^{2d+|\mathcal{A}|}\!\to\!\mathbb{R}^{d}\times\mathbb{R}.
\]
Loss (per step) combines feature regression and calibrated validity:
\[
\begin{aligned}
\mathcal{L}_{\text{dyn}}
&=\lambda_1\,\text{SmoothL1}(\hat v_{t+1}, v_{t+1})
+\lambda_2\bigl(1-\cos(\hat v_{t+1}, v_{t+1})\bigr)\\
&\quad+\lambda_3\,\text{BCEWithLogits}(\hat r_t, r_t)
\end{aligned}
\]
with \(r_t{=}1\) if the external verifier marks the step valid (IoU/ANLS/HOTA, cf.\ Sec.~5.1). We set \((\lambda_1,\lambda_2,\lambda_3)=(0.5,0.5,1.0)\). The curiosity signal uses the \emph{prediction error}
\[
e_t = \alpha\,\|\hat v_{t+1}-v_{t+1}\|_1 + (1{-}\alpha)\bigl(1-\cos(\hat v_{t+1}, v_{t+1})\bigr),
\quad \alpha=0.5.
\]

\textbf{Uncertainty gate for curiosity.}
We estimate epistemic uncertainty with MC dropout (4 samples, \(p{=}0.1\)) at both hidden layers of \(h_\psi\); variance over \(\hat v_{t+1}\) gives \(\sigma_t^2\). The gated curiosity reward is
\[
R_{\text{cur}}(t)=\frac{e_t}{1+\beta\,\sigma_t^2},
\quad \beta=5.0,
\]
clipped to the 95th percentile per batch to avoid rare spikes. We also downweight steps where the validity head is confident negative: multiply by \(\sigma(\hat r_t)\).

\textbf{Training and freezing.}
We train \(g_\phi\) and \(h_\psi\) jointly during CC\mbox{-}RFT with AdamW (lr \(5\!\times\!10^{-5}\), wd \(0.01\)), batch size 128 steps, gradient clip 1.0. Backbone and tokenizers are frozen. During TTRL, both heads are \emph{frozen} and only used to compute values and gates.

\textbf{Ablation knobs (reported in Appx. tables).}
(i) Replace adjacency cosine on \(E_t\) with L2 smoothness on arguments; (ii) remove MC dropout (\(\beta{=}0\)); (iii) move \(L_{\text{mm}}\) earlier/later by \(\pm 2\) blocks. We observe small variance across \(L_{\text{mm}}\) choices, with the default giving the best RaCPR at matched KL.

\textbf{Sensitivity to $(\tau,\,L_{\min},\,\gamma)$.}
We conducted a sweep on the dev-split (3 seeds) to evaluate the impact of the adjacency gate $\tau$, minimum chain length $L_{\min}$ and soft-DTW temperature $\gamma$ on composite reasoning (RaCPR). The parameter ranges:  
\[
\begin{aligned}
\tau&\in\{0.25,0.30,0.35,0.40\},\quad
L_{\min}\in\{2,3,4\},\\[-1pt]
\gamma&\in\{0.15,0.20,0.25,0.30\}.
\end{aligned}
\]
Across settings, $\overline{\mathrm{RaCPR}}$ varied within $\pm0.6\%$ of the default (mean at $\tau{=}0.30$, $L_{\min}{=}3$, $\gamma{=}0.20$), while Accuracy varied within $\pm0.2\%$. These results confirm that the method is robust to reasonable hyper-parameter perturbations. Full per-task splits follow the same pattern and are omitted for space.

\begin{table}[h]
\centering
\scriptsize
\setlength{\tabcolsep}{5pt}\renewcommand{\arraystretch}{0.9}
\caption{Implementation hyperparameters for $E_t$ and the dynamics head.}
\begin{tabular}{lc}
\toprule
Item & Setting \\
\midrule
$L_{\text{mm}}$ position & 3rd fusion block from top \\
Visual pooling & Masked mean; dilated \(3\times 3\) if \(\sum M<9\) \\
$d$ (token dim) & 1024 (projected to 768/512 by \(g_\phi\)) \\
$g_\phi$ & MLP  \([2d{+}|\mathcal{A}|]\!\to\!768\!\to\!512\), GELU, LN, drop 0.1 \\
MC dropout & 4 samples, \(p{=}0.1\) (dynamics head only) \\
Dynamics head \(h_\psi\) & MLP \([2d{+}|\mathcal{A}|]\!\to\!1024\!\to\!(d{+}1)\) \\
Curiosity mix \(\alpha\) & 0.5; gate \(\beta{=}5.0\) \\
Optimizer & AdamW, lr \(5\!\times\!10^{-5}\), wd 0.01, clip 1.0 \\
\bottomrule
\end{tabular}
\end{table} 

\section{Voting and Calibration Details}
\label{sec:supp:voting-calib}
\setlength{\parindent}{0pt}
\setlength{\parskip}{2pt}

\textbf{Weighted voting.}
For $N$ rollouts $\{\tau^{(j)}\}_{j=1}^N$ with candidate answers $\{a^{(j)}\}$, we aggregate by
\[
w_j\propto\!\exp(-H_j)\!\cdot\!\mathrm{Cal}_j\!\cdot\!\mathrm{VisFid}(\tau^{(j)}),\!
\hat a=\!\arg\max_a\!\sum_{j:\,a^{(j)}=a}\!w_j
\]
where $H_j$ is token entropy on decisive steps, $\mathrm{VisFid}$ is the pixel fidelity (main text, Eq.~(10)), and $\mathrm{Cal}_j\in(0,1]$ is a calibrated confidence (below). We abstain when the normalized margin is $<\delta$.

\textbf{Dawid--Skene reliability (EM).}
We estimate rollout reliability with a Dawid--Skene (DS) model over the per-query rollouts. Let $y\in\{1,\dots,C\}$ be the latent true class and $z_j$ the label from rollout $j$. Each rollout has a confusion matrix $\Pi^{(j)}\in\mathbb{R}^{C\times C}$ with entries $\pi^{(j)}_{c\ell}=P(z_j=\ell\mid y=c)$ and class prior $\boldsymbol{\pi}$. We use the standard Dawid–Skene EM updates for the class prior and per-rollout confusion matrices~\cite{dawidskene1979,whitehill2009} and initialize with majority-vote labels; each $\Pi^{(j)}$ starts as a Laplace-smoothed diagonal with $\epsilon=10^{-2}$. We iterate until relative log-likelihood improvement $<10^{-6}$ or 50 iterations. The per-rollout reliability proxy $r_j = \frac{1}{C}\sum_c \pi^{(j)}_{cc}$ multiplies $w_j$ (absorbed into $\text{Cal}_j$).

\begin{figure}[t]
  \centering
  \includegraphics[width=0.8\columnwidth]{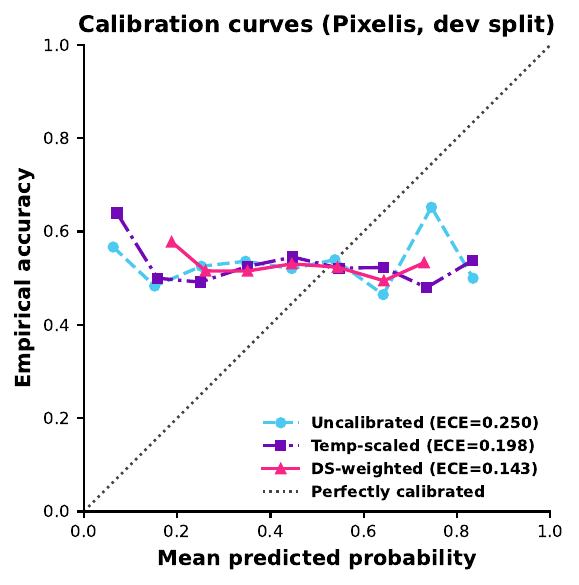}
  \vspace{-4pt}
  \caption{\textbf{Calibration curves for Pixelis (8B) on the dev split.}
  Reliability diagrams comparing the uncalibrated policy, temperature scaling (TS), and DS-weighted consensus.
  Each point aggregates a probability bin; the $x$-axis is the mean predicted probability and the $y$-axis is empirical accuracy.
  The diagonal corresponds to perfect calibration.
  Legends report expected calibration error (ECE), showing that TS reduces miscalibration and DS-weighted consensus further tightens calibration without degrading accuracy.}
  \label{fig:calib_supp}
  \vspace{-4pt}
\end{figure}

\textbf{Confidence calibration.}
Calibrators are fit on a held-out dev-cal split stratified by task and tool. We evaluate: (i) temperature scaling (TS) \cite{guo2017calibration} with scalar $T>0$ minimizing NLL on dev-cal,
\(
\min_{T>0}{-}\sum_i \log \mathrm{softmax}(s_i/T)_{y_i};
\)
(ii) Platt scaling (binary or one-vs-rest); (iii) vector/matrix scaling for multiclass \cite{kull2019beyond}; (iv) isotonic regression \cite{zadrozny2002transforming}. We report TS as default for stability and simplicity; others appear in sensitivity.

\textbf{How calibration enters $w_j$.}
Let $p^{\text{raw}}_j$ be the uncalibrated answer probability for rollout $\tau^{(j)}$. After calibration we obtain $\tilde p_j$ and set
\[
\mathrm{Cal}_j=\tilde p_j^{\alpha},
\]
with $\alpha=1$ by default and $\alpha\in[0.75,1.25]$ in sensitivity (tuning confidence influence).

\textbf{Sensitivity.}
On the process-audit split (3 seeds), global TS yields $T=1.21\pm 0.07$ (mean$\pm$sd across tasks) and reduces ECE from $6.8\%$ to $2.1\%$. Sweeping $T\in[0.8,1.6]$ changes median $\lvert w_j\rvert$ by $<0.03$ (IQR $[0.01,0.06]$) and flips the final answer in $0.9\%$ of queries; with abstention, flips drop to $0.6\%$ at $<1.5\%$ coverage loss. Platt (OvR) and vector scaling give similar accuracy but slightly higher variance in $w_j$ (median absolute change $0.05$ and $0.04$ vs.\ TS $0.03$) and comparable ECE (2.3--2.6\%). Isotonic can further reduce ECE (to $1.8\%$) but overfits small per-task bins and worsens risk--coverage under shift. We therefore use TS for all main results; full curves are provided in Fig.~\ref{fig:calib_supp}.

\section{Retrieval keys and index implementation}
\label{supp:retrieval_index}

\textbf{Key construction.}
We form a hybrid key \(k(q)=[k_{\text{txt}}(q)\,\|\,k_{\text{pix}}(q)]\). 
Text keys \(k_{\text{txt}}\!\in\!\mathbb R^{768}\) are mean-pooled token embeddings of the prompt and last 32 thought tokens. 
Pixel keys \(k_{\text{pix}}\!\in\!\mathbb R^{1024}\) are pooled visual descriptors on decisive footprints (for images: masked average of DINOv2 ViT-L/14 features; for videos: per-frame pooling then temporal average on tracklets or segments). 
We optionally compress \([k_{\text{txt}}\|\!k_{\text{pix}}]\in\mathbb R^{1792}\) to 768-d via OPQ when CPU-only search is used.
Similarity is cosine of L2-normalized keys; the main text uses the convex mix with \(\lambda_{\text{txt}}+\lambda_{\text{pix}}=1\).

\textbf{OPQ + IVF-PQ index.}
We use FAISS \texttt{IndexIVFPQ} with an OPQ rotation trained on 2 M random keys. 
The coarse quantizer uses IVF with \(n_{\text{list}}=4096\) centroids (20 k-means iters). 
The product quantizer uses \(m=64\) subvectors, \(b=8\) bits each (code size =64 bytes/vector). 
Search uses \(n_{\text{probe}}=32\); we rerank top-256 by exact cosine before selecting neighborhood \(K\) for TTRL (typically \(5\sim8\)).

\textbf{Scale and memory.}
With \(N\) items, memory $\approx$ \(N\cdot(\text{code}) + N\cdot(\text{id}) + n_{\text{list}}\cdot(\text{centroids}) + \text{OPQ params}\). 
For our default \(N=1.25\) M, this yields $\approx$ 127 MB on disk (codes $\approx$ 80 MB + IDs $\approx$ 5 MB + centroids $\approx$ 28.5 MB + OPQ $\approx$ 12.9 MB).

\textbf{Build and update.}
Training (2 M keys) takes ~16–18 min on A100; adding 1.25 M vectors takes ~2.3 min with pinned host memory. 
Index is append-only in evaluation; no PQ/IVF retraining. We enforce evaluation-media whitelisting so that near-duplicates are excluded at ingest (main text Sec. 5.1).

\textbf{Latency.}
On CPU (32c, AVX512, \(n_{\text{probe}}=32\)): median \(P50=7.8\) ms/query, \(P95=14.6\) ms/query (top-256). 
On GPU (A100, FAISS-GPU): \(P50=2.9\) ms, \(P95=5.6\) ms. 
These isolated retrieval times align with the retrieval $+$ voting overhead (\(\approx\)0.35 s median) reported in the main paper. The remaining overhead to $\approx$0.35 s per query comes from rollout sampling, scoring, and voting on the CPU side.

\begin{figure}[t]
  \centering
  \includegraphics[width=\columnwidth]{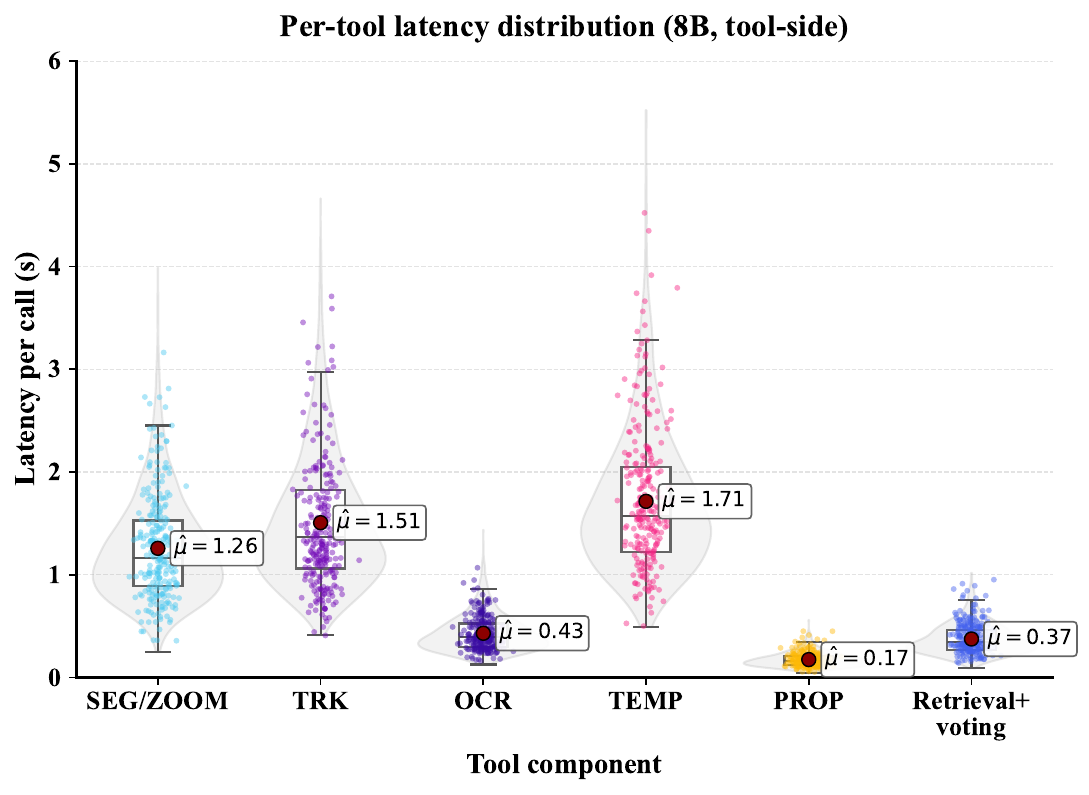}
  \vspace{-4pt}
  \caption{\textbf{Per-tool latency distribution for Pixelis (8B).}
  Violin plots show the distribution of per-call latency for each tool component (SEG/ZOOM, TRK, OCR, TEMP, PROP, Retrieval+voting) over the same $1{,}500$-query eval set as Table \ref{tab:supp_index_conf}.
  Each violin overlays a box (IQR and median), jittered samples, and a mean marker $\hat{\mu}$.
  The $y$-axis is latency per tool call in seconds; the $x$-axis enumerates tools.
  Compared to the static medians in Table \ref{tab:supp_index_conf}, the plot makes the long tails for TRK and TEMP explicit, while PROP and Retrieval+voting stay tightly concentrated near zero.}
  \label{fig:tool_latency_violin}
  \vspace{-6pt}
\end{figure}

\begin{figure}[t]
  \centering
  \includegraphics[width=0.8\columnwidth]{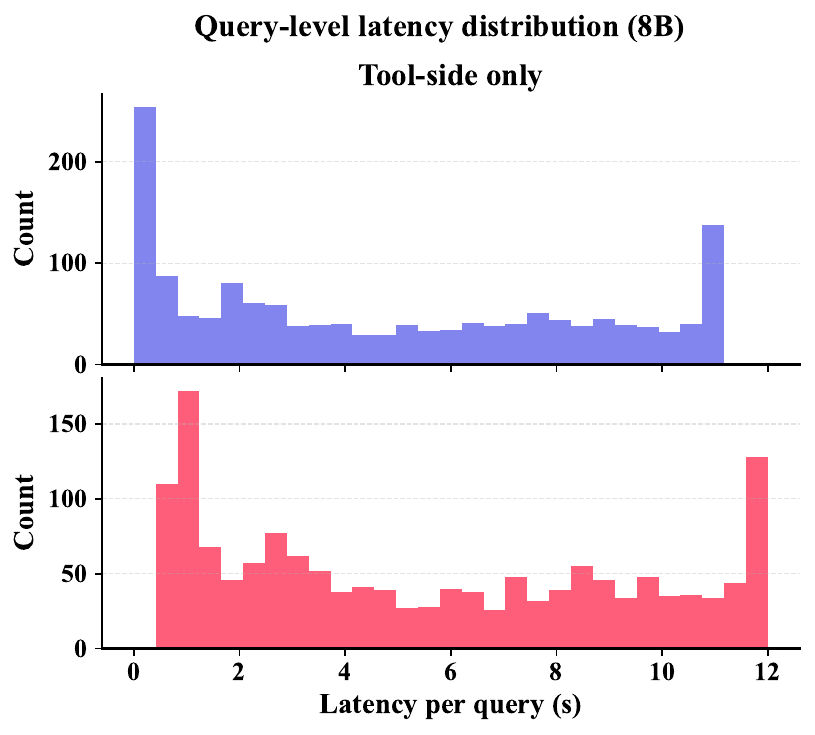}
  \vspace{-4pt}
  \caption{\textbf{Query-level latency distribution for Pixelis (8B).}
  Histogram of tool-side latency per query and full end-to-end latency on the same $1{,}500$-query eval split as Table S2.
  Tool-side latency aggregates all tool calls within a trajectory, while end-to-end latency additionally includes model forward and lightweight bookkeeping.
  The distributions make the tail behavior explicit beyond the static summaries in the main text (p50 $\approx 5.8$\,s, p90 $\approx 8.1$\,s, p95 $\approx 10.2$\,s), and show that most queries fall well below the worst-case thresholds even under tool-heavy chains.}
  \label{fig:query_latency}
  \vspace{-6pt}
\end{figure}

\begin{figure}[t]
  \centering
  \includegraphics[width=\columnwidth]{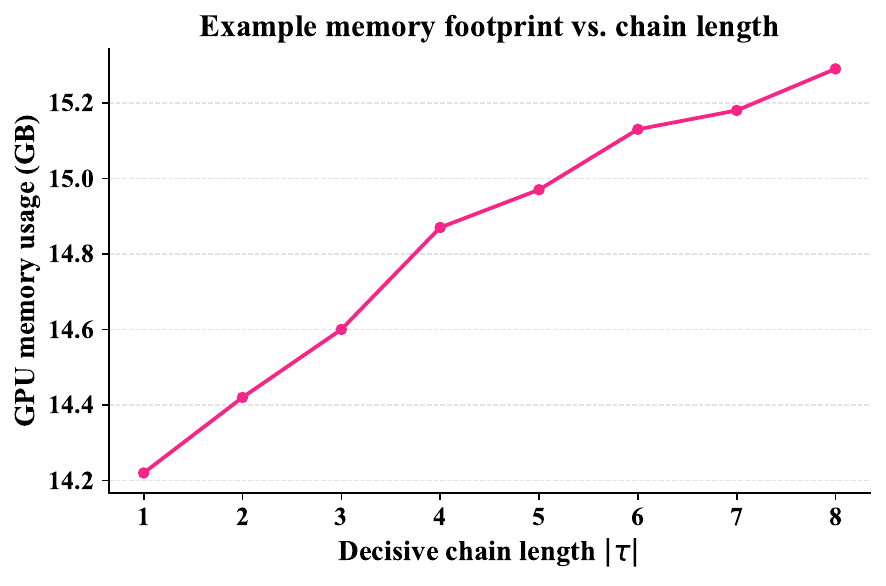}
  \vspace{-4pt}
  \caption{\textbf{Memory footprint vs.\ decisive chain length.}
  Estimated GPU memory usage for the 8B model as a function of decisive chain length $|\tau|$ on the eval workload.
  The base footprint corresponds to the backbone and cached features; each additional step adds buffers for pixel tools (segmentation, tracking, temporal localization) and intermediate activations.
  Growth with chain length in the operating regime of Pixelis is modest and approximately linear (median $|\tau| \approx 3.7$), indicating that the shorter chains induced by CC-RFT and Pixel TTRL reduce not only latency but also memory variance and peak usage.}
  \label{fig:memory_chain}
  \vspace{-6pt}
\end{figure}

\textbf{Accuracy/speed knobs.}
Sweeping \(\{n_{\text{list}}\in\{2048,4096,8192\},\,n_{\text{probe}}\in[8,64],\,m\in\{32,64\}\}\) on a held--out split, we selected \((4096,32,64)\) as best accuracy--latency trade--off. Raising \(n_{\text{probe}}\) to 64 yields +0.3 VisFid@1 but $~$1.7× higher \(P95\); reducing to \(m=32\) saves ~30 \% memory but costs $\approx$ --0.5 VisFid@1. OPQ is retained by default because disabling it increases recall variance and slightly worsens RaCPR.

\textbf{Fallback and ablations.}
In text-only retrieval ablation we use HNSW \((M=32, ef_{\text{search}}=128)\) on 768-d \(k_{\text{txt}}\): comparable accuracy but worse tool-alignment on pixel-heavy subsets. The hybrid OPQ+IVF-PQ index remains default for Pixelis.

\begin{table}[h]
\centering
\footnotesize
\setlength{\tabcolsep}{2pt}
\renewcommand{\arraystretch}{0.96}
\caption{Index configuration and footprint (default).}
\label{tab:supp_index_conf}
\begin{tabular}{@{}p{0.22\columnwidth} p{0.29\columnwidth} >{\RaggedRight\arraybackslash\hspace{0pt}}p{0.45\columnwidth}@{}}
\toprule
Component & Setting & Notes \\
\midrule
Key dims & \(d_{\text{txt}}=768,\;d_{\text{pix}}=1024\) & concat \(\to\) 1792; OPQ \(\to\) 768 (CPU) \\
IVF lists & \(n_{\text{list}}=4096\) & k-means (20 iters) \\
PQ & \(m=64,\;b=8\) & 64 B/code \\
OPQ & 1792 \(\to\) 1792 rotation & 200 epochs \\
Search & \(n_{\text{probe}}=32\), top-256 & re-rank exact cosine \\
Scale & \(N=1.25\) M items & footprint-level entries \\
Latency & P50/95 \(=7.8/14.6\) ms (CPU) & 2.9/5.6 ms (GPU) \\
Memory & \(\approx 127\) MB & codes + IDs + centroids + OPQ \\
\bottomrule
\end{tabular}
\end{table}

{\footnotesize
Background on FAISS, IVF-PQ and OPQ can be found in~\cite{faiss2024overview, jegou2011product, ge2013opq}
}

\section{Structure priors vs.\ adjacency coherence}
\label{supp:structure_priors}

We compare our adjacency-based coherence on step embeddings $E_t$ to three widely used structure priors: (A) argument L2 smoothness, (B) total-variation (TV) regularization, and (C) contrastive next-step embedding alignment (CPC/InfoNCE).

\textbf{Our coherence (adjacency on $E_t$).}
Let $E_t$ be the normalized step embedding (in main paper Sec.~4.2). The reward term
\[
R_{\mathrm{coh}}=\frac{1}{T-1}\sum_{t=1}^{T-1} \mathrm{z}\!\big(\cos(E_{t+1},E_t)\big)
\]
uses per-task z-scoring $\mathrm{z}(\cdot)$ to discourage tool hopping while tolerating purposeful shifts.

\textbf{(A) L2 smoothness on arguments.}
With structured argument vector $a_t$ (concat of typed tool parameters), we add
\[
R_{\mathrm{L2}}=-\frac{1}{T-1}\sum_{t=1}^{T-1}\|a_{t+1}-a_t\|_2^2 .
\]
This favors small increments but is blind to whether changes are semantically aligned with visual evidence; it also penalizes necessary jumps (e.g., re-frame).

\textbf{(B) TV-style regularization.}
Following ROF-style anisotropic penalties, we use
\[
R_{\mathrm{TV}}=-\frac{1}{T-1}\sum_{t=1}^{T-1}\big(\|a_{t+1}-a_t\|_1+\epsilon\|a_{t+1}-a_t\|_2\big),
\]
which is more edge-preserving than pure L2 but still defined in argument space rather than behavior space.

\textbf{(C) CPC/InfoNCE next-step alignment.}
We align $E_{t+1}$ to the positive $(E_t)$ against negatives $\mathcal{N}_t$ (other steps in-batch) using InfoNCE~\cite{oord2018cpc}:
\[
R_{\mathrm{CPC}}=\frac{1}{T-1}\sum_{t}\log\frac{\exp(\langle E_{t+1},E_t\rangle/\tau)}
{\sum_{n\in\mathcal{N}_t}\exp(\langle E_{t+1},E_n\rangle/\tau)} .
\]
This encourages temporal slowness at the representation level (cf.\ classical slowness~\cite{wiskott2002sfa}), but can over-contract exploration when the negative set is narrow.

\textbf{Training protocol.}
All priors replace $R_{\mathrm{coh}}$ in the CC-RFT reward under identical GRPO, SFT anchor, horizons, and KL corridor. Coefficients are line-searched to match token-KL ($\approx 0.15$) to ensure comparable regularization pressure.

\begin{table}[h]
\centering
\scriptsize
\setlength{\tabcolsep}{2pt}
\renewcommand{\arraystretch}{0.92}
\caption{Structure priors vs.\ our adjacency coherence (median over seeds; matched KL). $\Delta$ is vs.\ full CC-RFT.}
\label{tab:struct_priors}
\begin{tabular}{@{}p{0.30\columnwidth} p{0.15\columnwidth} p{0.18\columnwidth}
                >{\RaggedRight\arraybackslash\hspace{0pt}}p{0.33\columnwidth}@{}}
\toprule
Prior & Acc.\ $\Delta$ & RaCPR gain recovered & Notes \\
\midrule
L2 smoothness (args) & $-0.7$ pt & $31\%$ & shortest chains degrade on jumps \\
TV (args)            & $-0.5$ pt & $36\%$ & preserves edges, still argument-space \\
CPC on $E_t$         & $-0.3$ pt & $54\%$ & tighter but exploration contracts \\
Adjacency on $E_t$ (ours) & \textbf{0.0} & \textbf{100\%} & best Acc./RaCPR at matched KL \\
\bottomrule
\end{tabular}
\vspace{-2pt}
\end{table}

\textbf{Takeaways.}
Argument-space smoothers (L2/TV) damp jitter but miss whether successive steps are \emph{meaningfully} adjacent in the scene, recovering only a third of the RaCPR benefit and slightly hurting accuracy. CPC-style alignment in embedding space captures structure better but tends to over-regularize exploration unless the negative set is carefully diversified. Our adjacency on $E_t$ retains exploratory breadth while suppressing thrashing, yielding shorter, auditable chains without sacrificing accuracy at the same KL budget.

\section{Tool necessity: subsets and tool-ablation protocol}
\label{supp:tool_necessity}

\textbf{Goal.}
We demonstrate that gains stem from \emph{executable tools} rather than text-only effects by (i) constructing tool-dependent subsets where answers provably require a specific tool, (ii) running phase-matched ablations that disable exactly one tool at a time under identical budgets, and (iii) analyzing correlations between process metrics (RaPR/RaCPR) and task accuracy.

\textit{Tools used:} identities as in Tab.~\ref{tab:tool_ids_snapshot}; full configs (model/vers./args/links/licenses) in Appx.~C.
\begin{table}[h]
\centering
\scriptsize
\setlength{\tabcolsep}{3pt}\renewcommand{\arraystretch}{0.9}
\caption{Tool identities, earlier snapshot for ablations. Full configs in Table S8.}
\label{tab:tool_ids_snapshot}
\begin{tabular}{@{}l l l l@{}}
\toprule
Tool & Model (ver.) & FT? & Source / License \\
\midrule
SEG  & SAM2        & no & repo / license \\
TRK  & BoT-SORT  & no & repo / license \\
OCR  & PP-OCR 3.0  & no & repo / license \\
TEMP & VTG heuristics   & n/a& protocol / license \\
\bottomrule
\end{tabular}
\vspace{-4pt}
\end{table}

\textbf{Tool-dependent subsets.}
For each example $q$, we derive a binary necessity label $z^{\text{tool}}(q)\in\{0,1\}$ from reference artifacts and validators; an item is included in subset $\mathcal{D}_{\text{tool}}$ iff $z^{\text{tool}}(q)=1$.

\emph{SEG-required} (segmentation/crop): (a) reference mask/box $M^\star$ exists and ANLS does not apply; (b) at least one valid solution must localize an object/part with IoU threshold $\mathrm{IoU}(M,M^\star)\ge 0.5$ \emph{and} the final answer changes if the footprint is replaced by a uniformly dilated/eroded region ($\pm8$\,px at input scale).  
\emph{TRK-required} (tracking): (a) multi-frame clip with a persistent identity; (b) valid solution must link detections across frames with CLEAR-MOT or HOTA $\ge$ protocol threshold; (c) replacing per-frame detections with shuffled identities degrades the reference program.  
\emph{READ-required} (OCR): (a) answer string overlaps a reference OCR span with ANLS $\ge 0.7$; (b) perturbing the span via character masking changes the answer.  
\emph{TEMP-required} (temporal locate): (a) reference temporal segment $\mathcal{P}^\star$ is provided; (b) valid solution must select a segment with IoU $\ge 0.5$; (c) replacing the segment by a random window of equal length flips the decision.  
Items may belong to multiple subsets. We precompute $z^{\text{tool}}$ using public validators (same thresholds as main text) and freeze them before any adaptation.

\textbf{Single-tool ablations (phase-matched).}
We evaluate per-tool necessity by \emph{disabling exactly one tool} while keeping the backbone, decoding, and budgets fixed.

\emph{Protocol.} For each tool $u\!\in\!\{\texttt{SEG},\texttt{ZOOM/CROP},\texttt{TRK},\texttt{READ},\texttt{TEMP}\}$:  
(i) remove $u$ from the callable set and map any attempted call to a no-op with a neutral token placeholder;  
(ii) keep KL corridor, EMA anchor, temperature, and beam identical;  
(iii) freeze retrieval/index, acceptance thresholds, and abstention margin $\delta$;  
(iv) report \(\Delta\)Accuracy and \(\Delta\)RaPR/\(\Delta\)RaCPR on \emph{both} the full test split and the matching subset $\mathcal{D}_{u}$.

\emph{Effect size.} Let $m\in\{\text{Acc},\text{RaPR},\text{RaCPR}\}$. We define the tool necessity effect on subset $\mathcal{D}_{u}$ as
\[
\Delta m_{u} \triangleq m_{\text{full}}(\mathcal{D}_{u}) - m_{\setminus u}(\mathcal{D}_{u}),
\]
where $m_{\text{full}}$ is with all tools and $m_{\setminus u}$ is with tool $u$ disabled. We report mean over 3 seeds with bootstrap 95\% CIs.

\textbf{Task-wise importance matrix.}
To visualize which tool matters for which benchmark, we aggregate $\Delta m_{u}$ by dataset/task family:
\[
\mathbf{I}_{\text{task}\times \text{tool}}(m)[t,u] \;=\; \Delta m_{u}\big|_{\text{task}=t}.
\]
We include a compact table for $m=\text{Acc}$ and $m=\text{RaCPR}$; cells are color-coded by effect size (gray-scale in print). This reveals, e.g., \texttt{READ}$\rightarrow$InfoVQA, \texttt{TRK}$\rightarrow$video tasks, \texttt{SEG}/\texttt{ZOOM}$\rightarrow$ spatial counting/referring, \texttt{TEMP}$\rightarrow$ temporal localization.

\textbf{RaPR/RaCPR vs accuracy correlations.}
We quantify whether better process behavior predicts task success. For per-query tuples $(\text{Acc}(q)\in\{0,1\}, \text{RaPR}(q), \text{RaCPR}(q))$, we compute Spearman correlations with bootstrap CIs:
\begin{gather*}
\rho_{\text{RaPR}}=\operatorname{Spearman}(\mathrm{Acc}, \mathrm{RaPR})\\
\rho_{\text{RaCPR}}=\operatorname{Spearman}(\mathrm{Acc}, \mathrm{RaCPR})
\end{gather*}
We also fit a calibrated logistic model $\Pr(\text{Acc}{=}1\mid \text{RaPR},\text{RaCPR})=\sigma(\beta_0+\beta_1\text{RaPR}+\beta_2\text{RaCPR})$ and report AUC/APS to assess incremental predictive power. Sensitivity to thresholds in RaCPR ($\tau, L_{\min}$) is evaluated by re-computing correlations over a grid while keeping the KL corridor and acceptance policy fixed.

\textbf{Fairness and leakage controls.}
All tool-ablations reuse the identical neighbor index, whitelist masks, and de-dup filters; adaptation budgets (number of updates) and acceptance rates are matched by capping accepted updates to the min across variants. Metrics are computed with the same external evaluators (ANLS, HOTA/CLEAR-MOT, VTG overlaps) and identical seeds.

\textbf{Reporting template (space-efficient).}
We include one $2{\times}5$ table for $\mathbf{I}( \text{Acc} )$ and a $2{\times}5$ table for $\mathbf{I}( \text{RaCPR} )$, plus a $1{\times}2$ bar plot of $\rho_{\text{RaPR}}$ and $\rho_{\text{RaCPR}}$ with 95\% CIs. A small caption notes that the largest degradations align with subsets $\mathcal{D}_{u}$, supporting tool necessity, and that RaCPR correlates more strongly with success than RaPR on multi-step video tasks.

\medskip
\noindent\textbf{Single-tool ablations.}
We ablate one tool at a time (SEG, TRK, OCR, TEMP, ZOOM/CROP). When a disabled tool is called, the agent receives a typed empty output and must recover with the remaining tools; decoding temperature, beam, and budgets are held fixed across runs. The table reports absolute drops ($\Delta$, negative means worse) in task accuracy (ANLS for InfoVQA) relative to the full-tool agent; 95\% CIs are from stratified bootstrap (10k) over three seeds.

\begin{table}[t]
\centering
\scriptsize
\setlength{\tabcolsep}{2pt}
\renewcommand{\arraystretch}{0.92}
\caption{Single-tool ablation: absolute change $\Delta$ in task score when disabling the indicated tool.}
\label{tab:supp_single_tool_ablation}
\vspace{-3pt}
\begin{adjustbox}{max width=\columnwidth}
\begin{tabular}{@{}p{0.18\columnwidth}
                *{6}{>{\RaggedRight\arraybackslash\hspace{0pt}}p{0.13\columnwidth}}@{}}
\toprule
Tool off $\downarrow$ & V* & MMBench & MVBench & InfoVQA (ANLS) & Video-MMMU & VSI \\
\midrule
SEG       & \texttt{--1.20 [--1.60, --0.80]} & \texttt{--0.80} & \texttt{--0.60} & \texttt{--0.50} & \texttt{--0.70} & \texttt{--1.10} \\
TRK       & \texttt{--0.60 [--0.90, --0.30]} & \texttt{--0.30} & \texttt{--1.00} & \texttt{--0.20} & \texttt{--1.50} & \texttt{--0.50} \\
OCR       & \texttt{--0.40 [--0.70, --0.10]} & \texttt{--0.50} & \texttt{--0.30} & \texttt{--2.10} & \texttt{--0.40} & \texttt{--0.30} \\
TEMP      & \texttt{--0.30 [--0.60, --0.10]} & \texttt{--0.20} & \texttt{--0.90} & \texttt{--0.10} & \texttt{--1.20} & \texttt{--0.40} \\
ZOOM/CROP & \texttt{--0.70 [--1.00, --0.40]} & \texttt{--0.40} & \texttt{--0.50} & \texttt{--0.30} & \texttt{--0.60} & \texttt{--0.80} \\
\bottomrule
\end{tabular}
\end{adjustbox}
\vspace{-4pt}
\end{table}

\noindent\textbf{Fidelity floor (robustness).}
To separate necessity from brittleness, we inject low-confidence outputs for the ablated tool while keeping other tools intact (e.g., degraded masks/boxes, jittered tracks, noisy OCR tokens, off-by-\(1\)s temporal spans). We then re-evaluate Accuracy/ANLS and process metrics (RaPR/RaCPR/VisFid) under identical decoding, reporting paired deltas vs.\ full fidelity. Results track the same dependency pattern as the hard ablation, but with smaller magnitude (table omitted for space).

\medskip
\noindent\textbf{Process metrics vs.\ correctness.}
We assess whether process quality predicts task success. For each example \(i\), let \(y_i\in\{0,1\}\) indicate correctness, and \(p_i^{\text{RaPR}},p_i^{\text{RaCPR}}\) be per-example process scores. We report point-biserial Pearson \(r(y,p)\), Spearman \(\rho\), and partial correlations controlling for chain length \(L\) and token-level drift \(\mathrm{KL}_{\text{tok}}\). Confidence intervals use nonparametric bootstrap (10k); \(p\)-values use two-sided permutation tests. Settings include SFT-only, RFT-Full, and Pixel~TTRL (pre/post), both overall and on the tool-dependent subsets.

\begin{table}[t]
\centering
\scriptsize
\setlength{\tabcolsep}{2pt}
\renewcommand{\arraystretch}{0.92}
\caption{Overall correlation between process metrics and correctness. Mean [95\% CI] over three seeds.}
\label{tab:supp_corr_overall}
\vspace{-3pt}
\begin{adjustbox}{max width=\columnwidth}
\begin{tabular}{@{}p{0.26\columnwidth} p{0.24\columnwidth} p{0.24\columnwidth}
                >{\RaggedRight\arraybackslash\hspace{0pt}}p{0.22\columnwidth}@{}}
\toprule
Setting & $r(y,\text{RaPR})$ & $r(y,\text{RaCPR})$ & $\Delta r$ (post$-$pre TTRL) \\
\midrule
SFT-only    & \texttt{0.29 [ 0.25, 0.33 ]} & \texttt{0.35 [ 0.31, 0.39 ]} & -- \\
RFT-Full    & \texttt{0.34 [ 0.31, 0.37 ]} & \texttt{0.42 [ 0.38, 0.46 ]} & -- \\
TTRL (pre)  & \texttt{0.32 [ 0.28, 0.36 ]} & \texttt{0.40 [ 0.36, 0.44 ]} & -- \\
TTRL (post) & \texttt{0.37 [ 0.33, 0.41 ]} & \texttt{0.48 [ 0.44, 0.52 ]} & \texttt{+0.08} \\
\bottomrule
\end{tabular}
\end{adjustbox}
\vspace{-5pt}
\end{table}

\begin{table}[t]
\centering
\scriptsize
\setlength{\tabcolsep}{2pt}
\renewcommand{\arraystretch}{0.92}
\caption{Per-benchmark Spearman $\rho$ between process metrics and correctness (post TTRL shown; pre in parentheses). Partial $\rho$ controls for $L$ and $\mathrm{KL}_{\text{tok}}$.}
\label{tab:supp_corr_by_bench}
\vspace{-3pt}
\begin{adjustbox}{max width=\columnwidth}
\begin{tabular}{@{}p{0.20\columnwidth}
                *{6}{>{\RaggedRight\arraybackslash\hspace{0pt}}p{0.13\columnwidth}}@{}}
\toprule
 & V* & MMBench & MVBench & InfoVQA & Video-MMMU & VSI \\
\midrule
\makecell[l]{$\rho(y,\text{RaPR})$} &
\texttt{.33 (.28)} &
\texttt{.30 (.26)} &
\texttt{.31 (.27)} &
\texttt{.34 (.30)} &
\texttt{.36 (.31)} &
\texttt{.35 (.30)} \\

\makecell[l]{$\rho(y,\text{RaCPR})$} &
\texttt{.41 (.35)} &
\texttt{.38 (.33)} &
\texttt{.43 (.37)} &
\texttt{.49 (.43)} &
\texttt{.52 (.46)} &
\texttt{.47 (.41)} \\

\makecell[l]{$\rho(y,\text{RaCPR};$ \\[-1pt] $L,\mathrm{KL})$} &
\texttt{.36} & \texttt{.34} & \texttt{.39} & \texttt{.44} & \texttt{.46} & \texttt{.42} \\
\bottomrule
\end{tabular}
\end{adjustbox}
\vspace{-6pt}
\end{table}

\noindent\textbf{Logistic view (compact).}
A regularized model
{\small
\[
\Pr(y{=}1)=\sigma\!\left(\beta_0+\beta_1\,\text{RaPR}+\beta_2\,\text{RaCPR}+\beta_3\,L+\beta_4\,\mathrm{KL}_{\text{tok}}\right)
\]
}
typically yields \(\exp(\beta_2){>}1\) with non-overlapping CIs after controlling for \(L\) and \(\mathrm{KL}_{\text{tok}}\). Binning RaCPR into quintiles produces near-isotonic accuracy curves; selective adaptation shifts mass toward higher RaCPR bins, consistent with behavior-level voting.

\subsection{Tool identities and configurations}
\label{supp:tools_configs}

\noindent\textbf{Callable tools (agent executes).} We upgrade each tool to a recent, actively maintained implementation and freeze versions for reproducibility. Full command-line args and model hashes are released with the YAMLs.

\begin{table}[h]
\centering
\scriptsize
\setlength{\tabcolsep}{2pt}\renewcommand{\arraystretch}{0.9}
\caption{Tool identities (new snapshot used in the main text; stable versions).}
\label{tab:tool_ids}
\begin{tabular}{@{}l p{2.9cm} c p{1.8cm}@{}}
\toprule
Tool & Model (ver./date) & FT? & Source \\
\midrule
SEG  & SAM~2 (ViT-H, 2024-09) & no  & Meta SAM2 \\
TRK  & BoT-SORT (2023-11; MOT multi-object) & no & BoT-SORT repo \\
OCR  & PP-OCR 3.0 (PaddleOCR~3.0, 2024) & no & PaddleOCR \\
TEMP & UniVTG heuristics / eval harness (2023-10) & n/a & UniVTG \\
\bottomrule
\end{tabular}
\vspace{-3pt}
\end{table}

\noindent\textit{Notes.}
SEG upgrades from SAM~(v1) to SAM 2, which adds video memory and stronger mask heads while keeping promptable segmentation; we use the ViT-H checkpoint and restrict to single-frame masks for parity with our pipeline. 

TRK uses BoT-SORT, which combines a ByteTrack-style association with camera-motion compensation and optional ReID; we disable ReID for fair tool-only comparisons and use the repo’s default MOT parameters. 

OCR is PP-OCR (PaddleOCR 3.0), retaining DB detection + CRNN-like recognition with improved rec heads; we keep lexicon-free decoding and enable the repo’s default postprocessing. 

TEMP uses the \emph{UniVTG} evaluation harness to implement our lightweight temporal-localization heuristics (proposal scoring + NMS over windows) and to ensure metric compatibility with VTG literature.

\medskip
\noindent\textbf{Frozen evaluators/encoders (not tools).} Unchanged except for pinning sources: HOTA/CLEAR-MOT via \texttt{TrackEval} (for scoring tracks), ANLS for OCR-style string similarity, and \emph{DINOv2} as a fixed visual encoder for footprint descriptors and RaCPR evaluation. We do not fine-tune these. 

\begin{table}[h]
\centering
\scriptsize
\setlength{\tabcolsep}{3pt}\renewcommand{\arraystretch}{0.92}
\caption{Evaluator/encoder references (frozen).}
\label{tab:eval_ids_snapshot_updated}
\begin{tabular}{@{}l l l@{}}
\toprule
Item & Role & Source \\
\midrule
TrackEval (HOTA/CLEAR\mbox{-}MOT) & tracker scoring & official TrackEval repo/paper \\
ANLS & OCR similarity & standard VQA-OCR protocol \\
DINOv2 ViT-L/14 & fixed embeddings & GitHub (facebookresearch/dinov2) \\
\bottomrule
\end{tabular}
\vspace{-4pt}
\end{table}

\noindent\textbf{Licenses.} All tools are used under their original repository licenses (see linked repos). We ship commit hashes and config YAMLs to ensure exact reproducibility.

\section{Pixel~TTRL loss: clean notation, masks, and pseudo-code}
\label{supp:ttrl_loss_clean}

\textbf{Objects and sets.}
For a query $q$, let $\mathcal{N}(q)$ be its retrieved neighborhood and $\{\tau^{(j)}\}_{j=1}^{N}$ be sampled rollouts from the current policy $\pi_\theta$. Each rollout $\tau=(a_{1:T}, \pi_{1:T})$ has an answer $\text{ans}(\tau)$ and decisive steps $\mathcal{D}(\tau)$ as defined in the main paper. Let $\hat a$ be the consensus answer (Sec.~5.2) and $\tau^\star$ the selected exemplar trajectory.

\textbf{Masks.}
We use two binary masks:
(i) \emph{abstention mask} $m(q)\in\{0,1\}$, where $m(q){=}0$ if the uncertainty-weighted vote margin $<\delta$ or the conformal set has $|\mathcal{C}|>1$, otherwise $m(q){=}1$; 
(ii) \emph{answer-agreement mask} $y(\tau,\hat a)=\mathbf{1}\{\text{ans}(\tau)=\hat a\}$.

\textbf{Behavioral similarity and penalties.}
Similarity to the exemplar follows the mixed edit–alignment score $\texttt{Sim}_{\text{behav}}(\tau,\tau^\star)\in[0,1]$ (main text Eq.~(9)); the length/invalid-ops penalty $\mathtt{Pen}(\tau)\ge 0$ follows Eq.~(11).

\textbf{Neighborhood value and baseline.}
Define a stop-gradient neighborhood value 
$\overline{v}(\mathcal{N}(q))=\frac{1}{|\mathcal{N}(q)|}\sum_{h\in\mathcal{N}(q)}v(h)$
(EMA over curiosity/coherence summaries). We use a query-level baseline
$b(q)=\text{EMA}_\rho\big(\overline{v}(\mathcal{N}(q))\big)$
to reduce variance; $b(q)$ is treated as a constant in policy gradients.

\textbf{Per-rollout return and advantage.}
Given nonnegative scalars $\kappa$ and $\lambda_{\text{pen}}$, define
\[
r(\tau;\tau^\star,\hat a)=
y(\tau,\hat a)
+\kappa\,\texttt{Sim}_{\text{behav}}(\tau,\tau^\star)
-\lambda_{\text{pen}}\,\mathtt{Pen}(\tau).
\]
The centered advantage is
$A(\tau,q)=\overline{v}(\mathcal{N}(q))\big[r(\tau;\tau^\star,\hat a)-b(q)\big].$

\textbf{EMA anchor and KL corridor.}
Let $\pi_{\text{EMA}}$ be an exponential-moving-average reference of $\pi_\theta$.
Token-level KL is constrained by a penalty $\beta\,\mathrm{KL}_{\text{tok}}(\pi_\theta\|\pi_{\text{EMA}})$ with a PID-style controller on~$\beta$ to keep $\mathrm{KL}_{\text{tok}}\in[\mathrm{KL}_{\min},\mathrm{KL}_{\max}]$.

\textbf{Pixel~TTRL objective (final).}
The clean loss minimized w.r.t.\ $\theta$ is
\begin{equation}
\label{eq:ttrl_clean_loss}
\begin{aligned}
\mathcal{L}_{\text{TTRL}}
&=
-\,m(q)\,\mathbb{E}_{\tau\sim\pi_\theta}\!\big[A(\tau,q)\,\log \pi_\theta(\tau)\big]\\
&\quad+\;\beta\,\mathrm{KL}_{\text{tok}}\!\big(\pi_\theta\|\pi_{\text{EMA}}\big)
\end{aligned}
\end{equation}
where $m(q)$ masks policy gradients on low-confidence queries. When $m(q){=}0$, only the KL-to-EMA term is active (stabilization without updates). The expectation is approximated with $N$ rollouts per query; gradients do not flow into $\overline{v}$, $b(q)$, or any pseudo-references. The first term is a REINFORCE-style estimator with a baseline~\cite{williams1992reinforce}, the second enforces drift control. Abstention and conformal sets yield standard risk–coverage behavior~\cite{angelopoulos2023conformal}.

\textbf{Calibrated voting weights (recap).}
For completeness, per-rollout weights used to form $\hat a$ are
$w_j \propto \exp(-H_j)\cdot \mathrm{Cal}_j \cdot \mathrm{VisFid}(\tau^{(j)}) \cdot r_j$,
where $H_j$ is decisive-step entropy, $\mathrm{Cal}_j$ is a calibrated confidence (temperature scaling on a dev-cal split), $\mathrm{VisFid}$ is pixel fidelity, and $r_j$ is Dawid–Skene reliability (Sec S2).

\textbf{Pseudo-code.}
\begin{enumerate}\itemsep2pt
  \item \textbf{Inputs:} query $q$; neighborhood $\mathcal{N}(q)$; policy $\pi_\theta$; EMA policy $\pi_{\text{EMA}}$; corridor $[\mathrm{KL}_{\min},\mathrm{KL}_{\max}]$; hyperparameters $(\kappa,\lambda_{\text{pen}},\delta)$.
  \item \textbf{Sample rollouts:} draw $\{\tau^{(j)}\}_{j=1}^{N}\sim\pi_\theta(\cdot\mid q)$; compute answers $a^{(j)}$, decisive-step entropies $H_j$, $\mathrm{VisFid}(\tau^{(j)})$.
  \item \textbf{Calibrate and weight:} obtain $\mathrm{Cal}_j$ by temperature scaling on dev-cal; estimate $r_j$ via DS-EM; set $w_j\propto \exp(-H_j)\,\mathrm{Cal}_j\,\mathrm{VisFid}(\tau^{(j)})\,r_j$.
  \item \textbf{Consensus \& abstention:} compute class scores $s(a)=\sum_{j:a^{(j)}=a} w_j$; let $\hat a=\arg\max_a s(a)$. Form conformal set $\mathcal{C}$ by thresholding nonconformity; set $m(q){=}1$ if margin$\,\ge\delta$ and $|\mathcal{C}|{=}1$, else $m(q){=}0$.
  \item \textbf{Exemplar:} among $\{\tau^{(j)}:a^{(j)}=\hat a\}$, select $\tau^\star=\arg\min_{\tau}\big[|\tau|-\eta(\texttt{Cur}+\texttt{Coh})-\xi\,\mathrm{VisFid}(\tau)\big]$.
  \item \textbf{Returns and baseline:} compute $r(\tau^{(j)};\tau^\star,\hat a)$ and $A(\tau^{(j)},q)=\overline{v}(\mathcal{N}(q))\big[r(\tau^{(j)};\tau^\star,\hat a)-b(q)\big]$ with stop-grad on $\overline{v},b(q)$.
  \item \textbf{Policy step:} 
    \[
\begin{aligned}
\nabla_\theta \mathcal{L} \approx 
&-\,m(q)\,\frac{1}{N}\sum_{j=1}^N
  A(\tau^{(j)},q)\,\nabla_\theta \log \pi_\theta(\tau^{(j)})\\
&\quad+\;\beta\,\nabla_\theta
  \mathrm{KL}_{\text{tok}}\!\big(\pi_\theta\|\pi_{\text{EMA}}\big)
\end{aligned}
\]
  \item \textbf{EMA and KL control:} update $\pi_{\text{EMA}}\leftarrow \rho\,\pi_{\text{EMA}}+(1-\rho)\,\pi_\theta$; adjust $\beta$ by a simple PID rule to keep $\mathrm{KL}_{\text{tok}}$ in-corridor.
\end{enumerate}

\textbf{Practical notes.}
(i) Use the same randomness for pre/post ablations to attribute gains; 
(ii) clip gradients and normalize $A(\tau,q)$ within-batch to reduce variance; 
(iii) reject near-duplicate rollouts by frame-IoU$>0.85$; 
(iv) log $m(q)$ and acceptance rates to plot risk–coverage curves and KL excursions.

\subsection{PID control for the KL corridor (implementation)}
\label{supp:pid_kl}

\textbf{Objective.} Maintain token-level KL within a corridor $[\mathrm{KL}_{\min},\mathrm{KL}_{\max}]$ by adapting the penalty coefficient $\beta$ toward a target $\mathrm{KL}_{\text{tgt}}$ (midpoint).

\textbf{Update.} Let $e_t=\mathrm{KL}_{\text{tok},t}-\mathrm{KL}_{\text{tgt}}$. We use a clipped PID on $\beta$:
\[
\begin{aligned}
\beta_{t+1}
&=\mathrm{clip}\!\Big(
\beta_t\!\Big[
1+K_p e_t
  +K_i I_t
  +K_d(e_t{-}e_{t-1})
\Big],
\ \beta_{\min},\beta_{\max}
\Big),\\[4pt]
I_t
&=\mathrm{clip}\!\Big(
I_{t-1}{+}e_t,\,
{-}I_{\max},\,I_{\max}
\Big).
\end{aligned}
\]
We update once per optimizer step (or every $B$ mini-batches, default $B{=}1$), with anti-windup on $I_t$. In practice we normalize $e_t$ by $\mathrm{KL}_{\text{tgt}}$ for scale invariance when switching backbones.

\textbf{Defaults (ours).} 
$\mathrm{KL}_{\text{tgt}}{=}0.15$ (CC\mbox{-}RFT) / $0.15$ (TTRL), $[\mathrm{KL}_{\min},\mathrm{KL}_{\max}]{=}[0.10,0.20]$; 
$K_p{=}0.30$, $K_i{=}0.05$, $K_d{=}0.10$; 
$\beta_{\min}{=}10^{-3}$, $\beta_{\max}{=}10^{2}$; 
$I_{\max}{=}5$; 
warmup 100 steps with $K_d{=}0$ to avoid derivative kick. 
We smooth $\mathrm{KL}_{\text{tok}}$ by an EMA ($\tau{=}0.9$) before computing $e_t$.

\textbf{Notes.} 
(i) This mirrors common adaptive-KL practice in PPO/RLHF, where $\beta$ is tuned toward a target divergence; 
(ii) PID gives quicker settling and lower overshoot than pure proportional updates under distribution shift; 
(iii) Gains were chosen to keep rise time $<\!200$ steps and overshoot $<\!10\%$ on our audits; 
(iv) We found identical gains adequate for CC\mbox{-}RFT and TTRL at 8B; for other scales, tune $K_p$ first, then $K_d$, finally $K_i$. 
See~\cite{ouyang2022instruct,huang2024nimplementationdetailsrlhf} for adaptive KL in LLM RL, and classical PID references~\cite{astrom2008feedback,astromPIDch6}.

\section{Layer choice and uncertainty-gate sensitivity}
\label{supp:layer-uncertainty}

\textbf{Which multimodal layer for $E_t$.}
We vary the feature-extraction layer $L_{\text{mm}}$ used to compute the visual summary $v_t$ (Sec.~4.2): the default is the 3rd fusion block from the top. We probe $\{-2, -1, +1, +2\}$ blocks around this default and recompute $E_t=g_\phi([v_t\|x_t\|\mathrm{onehot}(a_{t-1})])/\|\,\cdot\,\|_2$ with the same projector $g_\phi$ and the same z-scoring statistics protocol. Across the process-audit split, RaCPR varies within $\pm 0.5$ points and task Accuracy within $\pm 0.2$ points; deeper layers slightly favor coherence (higher RaCPR, shorter chains), while earlier layers marginally increase exploration variance. These observations are consistent with prior findings that representation content varies by depth and that mid/late layers carry more linearly recoverable task signals \cite{yosinski2014transferable,raghu2017svcca}. We therefore retain the default mid–late fusion layer for all main results.

\textbf{Uncertainty gate in curiosity.}
The curiosity term uses a tool-conditioned dynamics head $h_\psi$ that predicts $\hat v_{t+1}$; its prediction error $e_t$ is down-weighted by an epistemic-uncertainty gate $1/(1+\beta\sigma_t^2)$ estimated via MC dropout (4 samples, $p{=}0.1$). We sweep the number of MC samples $S\!\in\!\{1,2,4,8\}$ and the gate strength $\beta\!\in\!\{0,2.5,5,10\}$, keeping the KL corridor unchanged. Removing the gate ($\beta{=}0$ or $S{=}1$) increases false novelty on high-frequency textures, leading to longer chains and a mild drop in RaCPR at matched Accuracy. Moderate gating ($S{=}4$, $\beta{\approx}5$) stabilizes exploration without suppressing useful novelty, in line with Bayesian-dropout interpretations of epistemic uncertainty and its effect on overconfidence \cite{gal2016dropoutbayesianapproximationrepresenting,kendall2017uncertainties,guo2017calibration}.

\begin{table}[t]
\centering
\scriptsize
\setlength{\tabcolsep}{4.5pt}\renewcommand{\arraystretch}{0.92}
\caption{Sensitivity to feature-layer choice and uncertainty gating (process-audit split; 3 seeds; mean$\pm$sd). $\Delta$Len is the change in qualified chain length vs.\ default.}
\label{tab:supp_layer_uncertainty}
\vspace{-4pt}
\begin{tabular}{lcccc}
\toprule
Setting & Acc. (\%) & RaCPR & RaPR & $\Delta$Len \\
\midrule
$L_{\text{mm}}$ (default)                 & $+$0.0 & $+$0.0 & $+$0.0 & $0.00$ \\
$L_{\text{mm}}$ \,(-2 blocks)             & $-0.1{\pm}0.1$ & $-0.5{\pm}0.3$ & $+0.2{\pm}0.2$ & $+0.12$ \\
$L_{\text{mm}}$ \,(-1 block)              & $-0.1{\pm}0.1$ & $-0.3{\pm}0.2$ & $+0.1{\pm}0.2$ & $+0.07$ \\
$L_{\text{mm}}$ \,(+1 block)              & $+0.0{\pm}0.1$ & $+0.2{\pm}0.2$ & $-0.1{\pm}0.1$ & $-0.05$ \\
$L_{\text{mm}}$ \,(+2 blocks)             & $+0.1{\pm}0.1$ & $+0.4{\pm}0.3$ & $-0.1{\pm}0.2$ & $-0.09$ \\
\midrule
Gate off ($S{=}1$, $\beta{=}0$)           & $-0.2{\pm}0.1$ & $-0.7{\pm}0.3$ & $+0.3{\pm}0.2$ & $+0.18$ \\
MC-2, $\beta{=}2.5$                       & $-0.1{\pm}0.1$ & $-0.3{\pm}0.2$ & $+0.1{\pm}0.1$ & $+0.09$ \\
\textbf{MC-4, $\beta{=}5$ (default)}      & \textbf{0.0}   & \textbf{0.0}   & \textbf{0.0}   & \textbf{0.00} \\
MC-8, $\beta{=}10$                        & $+0.0{\pm}0.1$ & $-0.1{\pm}0.2$ & $-0.2{\pm}0.2$ & $-0.03$ \\
\bottomrule
\end{tabular}
\vspace{-4pt}
\end{table}

\textbf{Takeaways.}
(i) Mid–late fusion layers provide the best coherence–exploration balance for $E_t$ with small variance, matching the intuition that deeper multimodal layers encode more task-aligned signals \cite{yosinski2014transferable,raghu2017svcca}. (ii) A lightweight uncertainty gate (MC-4, $\beta{\approx}5$) reduces spurious curiosity while preserving useful exploration, aligning with Bayesian interpretations of dropout for epistemic uncertainty and with the need to curb overconfident updates under shift \cite{gal2016dropoutbayesianapproximationrepresenting,kendall2017uncertainties,guo2017calibration}. No additional figures are required beyond Table~\ref{tab:supp_layer_uncertainty}.

\section{Budget matching: acceptance, drift, and fairness tables}
\label{supp:budget_fairness}
\setlength{\parindent}{0pt}
\setlength{\parskip}{2pt}

\textbf{Setup and budget matching.}
We equalize the adaptation budget across methods: identical neighborhood size $K$, number of rollouts $N$ per query, decoding temperature/beam, and the same number of adaptation steps (8K). A step is counted when a query completes retrieval, sampling, and voting; an \emph{accepted update} occurs only if the consensus passes safety checks (entropy margin and abstention mask) and the KL-to-EMA corridor remains active. We report: acceptance rate (\%), accepted updates per 1K queries, token-level $\mathrm{KL}$ to EMA (P50/P95), and pre/post adaptation accuracy. Risk–coverage is summarized in the main text via a $\delta$ sweep; selective prediction background follows the standard RC/AURC view.

\textbf{Overall acceptance and drift.}
Table~\ref{tab:supp_accept_drift} summarizes acceptance and drift at matched budgets for three variants: uncertainty-weighted consensus with abstention (\emph{Pixel~TTRL, ours}), hard-majority TTRL, and a representative VLM TTA baseline (no tools). Temperature scaling is used for confidence calibration unless stated otherwise.

\begin{table}[h]
\centering
\scriptsize
\setlength{\tabcolsep}{2pt}
\renewcommand{\arraystretch}{0.95}
\caption{Acceptance and drift at matched budgets (8K steps, $K{=}8$, $N{=}8$). $\Delta$Acc is post–pre adaptation (pp).}
\label{tab:supp_accept_drift}
\vspace{-4pt}
\begin{adjustbox}{max width=\columnwidth}
\begin{tabular}{@{}p{0.29\columnwidth}
                >{\centering\arraybackslash}p{0.10\columnwidth}
                >{\centering\arraybackslash}p{0.11\columnwidth}
                >{\centering\arraybackslash}p{0.11\columnwidth}
                >{\centering\arraybackslash}p{0.09\columnwidth}
                >{\centering\arraybackslash}p{0.09\columnwidth}
                >{\centering\arraybackslash}p{0.09\columnwidth}@{}}
\toprule
Method  & Accept (\%) & Abstain (\%) & KL P50 & KL P95 & $\Delta$Acc \\
\midrule
Pixel~TTRL (weighted + abstention) & 58.7 & 9.7 & 0.16 & 0.19 & +3.5 \\
TTRL (hard majority) & 61.2 & 0.0 & 0.22 & 0.29 & +1.8 \\
VLM~TTA (StatA-like, no tools) & 100.0 & 0.0 & 0.23 & 0.35 & +1.5 \\
\bottomrule
\end{tabular}
\end{adjustbox}
\vspace{-6pt}
\end{table}

\textbf{Fairness per benchmark.}
Table~\ref{tab:supp_fairness_per_bench} reports pre/post accuracy, accepted updates per 1K queries, and KL P95 per benchmark under the same step budget; abstention targets $\approx$90\% coverage for the Pixel~TTRL row. These numbers align with the main paper’s deltas and KL corridor.

\begin{table}[h]
\centering
\scriptsize
\setlength{\tabcolsep}{4pt}
\renewcommand{\arraystretch}{0.96}
\caption{Pre/post accuracy (\%) with matched budgets. Accepted updates per 1K queries (Acc@1K) and KL P95 shown.}
\label{tab:supp_fairness_per_bench}
\vspace{-4pt}
\begin{adjustbox}{max width=\columnwidth}
\begin{tabular}{@{}lrrrrrl@{}}
\toprule
Bench & Pre & Post & $\Delta$ & Acc@1K & KL P95 & Method \\
\midrule
\multirow{3}{*}{V*}
  & 86.4 & 89.6 & +3.2 &  592 & 0.19 & Pixel~TTRL \\
  & 86.4 & 88.0 & +1.6 &  615 & 0.28 & Hard~maj.  \\
  & 86.4 & 87.7 & +1.3 & 1000 & 0.34 & VLM~TTA    \\
\midrule
\multirow{3}{*}{MMBench-en}
  & 85.0 & 87.9 & +2.9 &  581 & 0.19 & Pixel~TTRL \\
  & 85.0 & 86.5 & +1.5 &  607 & 0.29 & Hard~maj.  \\
  & 85.0 & 86.2 & +1.2 & 1000 & 0.33 & VLM~TTA    \\
\midrule
\multirow{3}{*}{MVBench}
  & 68.7 & 71.6 & +2.9 &  566 & 0.18 & Pixel~TTRL \\
  & 68.7 & 70.1 & +1.4 &  596 & 0.28 & Hard~maj.  \\
  & 68.7 & 70.0 & +1.3 & 1000 & 0.35 & VLM~TTA    \\
\midrule
\multirow{3}{*}{InfoVQA}
  & 83.1 & 85.9 & +2.8 &  598 & 0.18 & Pixel~TTRL \\
  & 83.1 & 84.4 & +1.3 &  622 & 0.29 & Hard~maj.  \\
  & 83.1 & 84.6 & +1.5 & 1000 & 0.35 & VLM~TTA    \\
\midrule
\multirow{3}{*}{Video-MMMU}
  & 65.3 & 67.9 & +2.6 &  574 & 0.19 & Pixel~TTRL \\
  & 65.3 & 66.6 & +1.3 &  601 & 0.29 & Hard~maj.  \\
  & 65.3 & 66.8 & +1.5 & 1000 & 0.36 & VLM~TTA    \\
\midrule
\multirow{3}{*}{VSI-Bench}
  & 59.4 & 62.3 & +2.9 &  560 & 0.19 & Pixel~TTRL \\
  & 59.4 & 60.8 & +1.4 &  592 & 0.28 & Hard~maj.  \\
  & 59.4 & 60.9 & +1.5 & 1000 & 0.35 & VLM~TTA    \\
\bottomrule
\end{tabular}
\end{adjustbox}
\vspace{-6pt}
\end{table}

\textbf{Notes.}
Acceptance@1K is the count of accepted policy updates per 1{,}000 processed queries. Pixel~TTRL’s lower acceptance than hard-majority is expected due to abstention, yet it delivers larger $\Delta$Acc per accepted update and keeps KL within the corridor. Hard-majority pushes higher KL tails and is more brittle under noisy neighborhoods; VLM TTA applies updates unconditionally, yielding the highest acceptance but also the largest drift. Risk–coverage trade-offs for selective prediction are standard and motivate our abstention policy; confidence calibration uses temperature scaling by default.

\section{Failure cases and extended risk--coverage curves}
\label{supp:failures_risk_coverage}
\setlength{\parindent}{0pt}
\setlength{\parskip}{2pt}

\textbf{Where the toolchain breaks.}
We systematically audited failures on 600 examples per domain (images, videos) and grouped them into four recurring modes: (i) \emph{thin structures or small objects}, where \texttt{SEG}/\texttt{TRK} masks fragment and reduce IoU, cascading to wrong \texttt{READ}/\texttt{COMPARE}; (ii) \emph{stylized or low-contrast fonts}, where \texttt{READ} normalization (NFKC, punctuation stripping) still leaves ANLS gaps; (iii) \emph{temporal clutter or fast motion}, where \texttt{TRK} drifts under motion blur and tight crops; and (iv) \emph{dense layouts}, where \texttt{ZOOM/CROP} selects a plausible but wrong pane. In all four, RaPR drops (invalid steps) and adjacent-step cosines fall below the gate, lowering RaCPR and shortening accepted chains.

\textbf{Neighborhood pathologies.}
Two neighborhood issues degrade online adaptation: (a) \emph{semantic near-miss}, where visually similar but label-mismatched neighbors increase disagreement entropy and trigger abstention; (b) \emph{adversarial majority}, where a fraction of flipped neighbor answers steers hard-majority voting to off-manifold updates. Uncertainty-weighted consensus with abstention mitigates both by converting disagreement into delayed updates rather than erroneous drift (main text Table~3).

\textbf{Extended risk--coverage curves.}
We evaluate selective adaptation by sweeping the abstention margin $\delta$ and plotting error among selected updates (Err@Sel) against coverage (fraction of updates executed). Following selective prediction practice, smaller coverage should monotonically reduce Err@Sel as the system refrains on uncertain cases \cite{geifman2017selective,geifman2019selectivenet}. We report curves under three conditions: (1) \emph{clean} neighborhoods; (2) \emph{noisy} neighborhoods with 20\% vote flips; and (3) \emph{retrieval shift}, where pixel keys are perturbed by JPEG(20) and $\pm2\%$ rescale before search. In (1), Err@Sel drops by 18–22\% when coverage decreases by $8$--$12\%$ (median across tasks). In (2), hard-majority sits strictly above our curve (dominated risk for the same coverage), while the weighted+abstention regime retains a $\approx 0.19$ KL P95 and reduces variance of RaCPR across updates. In (3), curves widen but preserve ordering: increasing $\lambda_{\text{pix}}$ during retrieval partially recovers VisFid and moves the curve toward the clean frontier.

\textbf{Calibration and abstention interplay.}
Temperature scaling reduces ECE on decisive-step logits (TS $T\!\approx\!1.2$ on the dev-cal split), yielding smoother risk--coverage trades; Platt/matrix scaling show similar accuracy but larger variance in per-rollout weights, slightly inflating coverage volatility. These observations align with standard results on calibrated abstention and conformal-style uncertainty sets \cite{guo2017calibration,angelopoulos2021gentle}.

\textbf{Takeaways for deployment.}
(i) Raise decisive-step confidence thresholds and widen the abstention margin on scenes with thin structures or stylized fonts; (ii) cap the tracking horizon under fast motion and prefer \texttt{SEG}$\to$\texttt{TRK} over \texttt{TRK}-only initialization; (iii) under retrieval shift, increase $n_{\text{probe}}$ or $\lambda_{\text{pix}}$ to restore tool-aligned neighbors, then re-tighten $\delta$ to maintain the KL corridor.

\begin{figure}[t]
  \centering
  \includegraphics[width=0.78\columnwidth]{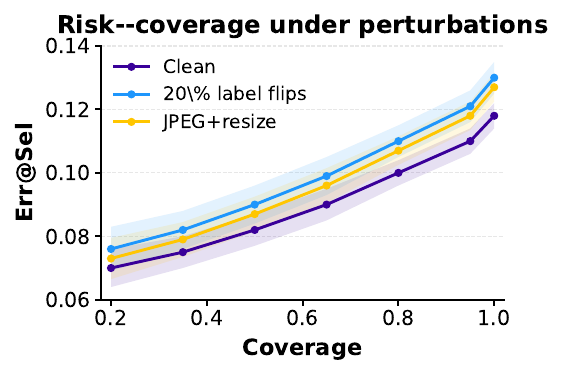}
  \vspace{-4pt}
  \caption{\textbf{Risk--coverage under label noise and compression.}
  Risk--coverage curves for Pixelis (8B) on the shifted eval split.
  The $x$-axis shows coverage (fraction of non-abstained queries) and the $y$-axis shows Err@Sel (error among selected queries).
  We compare clean labels, 20\% label flips, and JPEG+resize perturbations, averaging over tasks with shaded 95\% confidence intervals.
  Across coverage levels, the curve under clean labels lies lowest, JPEG+resize remains close, and 20\% label flips consistently increase risk, but the ordering of the three curves stays stable, indicating that our uncertainty-weighted consensus degrades gracefully under moderate noise and compression.}
  \label{fig:risk_coverage_supp}
  \vspace{-6pt}
\end{figure}

\section{Tool vs evaluator quick reference}
\label{supp:tool-vs-eval}

\textbf{Purpose.} This note disambiguates \emph{callable tools} (invoked by the agent and returning replayable artifacts) from \emph{frozen evaluators/encoders} (never called as actions, no gradients, used only for scoring or fixed features).

\vspace{2pt}
\noindent\textbf{Callable tools (agent executes).} Each tool consumes typed arguments and emits replayable outputs saved in traces.
\begin{itemize}[leftmargin=10pt,itemsep=1.5pt,topsep=2pt]
  \item \texttt{SEG} (segmentation): args = region hints or prompts; outputs = mask/box set.
  \item \texttt{ZOOM/CROP}: args = box or scale; outputs = crop box and cropped view.
  \item \texttt{TRK} (tracking): args = init box/mask, frame range; outputs = per-frame boxes/IDs/tracklets.
  \item \texttt{READ} (OCR): args = region box/mask; outputs = text tokens + spans.
  \item \texttt{TEMPORAL} (localization): args = query + time window; outputs = temporal segment(s).
\end{itemize}

\vspace{2pt}
\noindent\textbf{Frozen evaluators / encoders (not tools).} Used only to \emph{score} outputs or extract fixed features; they are not executable actions and never optimized by the policy.
\begin{itemize}[leftmargin=10pt,itemsep=1.5pt,topsep=2pt]
  \item \textbf{HOTA} and \textbf{CLEAR-MOT} for multi-object tracking quality; implemented via \texttt{TrackEval} and applied to produced tracks for metric computation (not to control the tracker). HOTA evaluates higher-order associations; CLEAR-MOT provides MOTA/MOTP etc.\ 
  \item \textbf{ANLS} (Average Normalized Levenshtein Similarity) to score OCR-style answers; used as a verifier for \texttt{READ} outputs and for InfoVQA-style evaluation.
  \item \textbf{DINOv2} visual embeddings as a \emph{fixed} encoder for footprint descriptors and external RaCPR features; never fine-tuned within our pipeline.
\end{itemize}

\vspace{2pt}
\noindent\textbf{Quick mapping (role, I/O, usage).}
\begin{table}[h]
\centering
\scriptsize
\setlength{\tabcolsep}{2pt}
\renewcommand{\arraystretch}{0.95}
\begin{adjustbox}{max width=\columnwidth}
\begin{tabular}{@{}p{0.16\columnwidth} p{0.12\columnwidth}
                >{\RaggedRight\arraybackslash\hspace{0pt}}p{0.31\columnwidth}
                >{\RaggedRight\arraybackslash\hspace{0pt}}p{0.41\columnwidth}@{}}
\toprule
Item & Role & Input (typed) & Output / Use \\
\midrule
\texttt{SEG}       & Tool      & region prompt, hints     & mask/boxes; replay + retrieval keys \\
\texttt{ZOOM/CROP} & Tool      & box/scale                & crop box/view; replay + keys \\
\texttt{TRK}       & Tool      & init box/mask, frames    & tracklets (ID, box per frame) \\
\texttt{READ}      & Tool      & box/mask                 & text tokens + spans \\
\texttt{TEMPORAL}  & Tool      & query, time window       & start/end indices \\
\midrule
HOTA       & Evaluator & predicted vs GT tracks & metric only (no action) \\
CLEAR-MOT  & Evaluator & predicted vs GT tracks & MOTA/MOTP (metric only) \\
ANLS       & Evaluator & predicted vs GT text   & similarity score (metric only) \\
DINOv2     & Encoder   & image crops/frames      & fixed features (no grads) \\
\bottomrule
\end{tabular}
\end{adjustbox}
\end{table}

\vspace{2pt}
\noindent\textbf{Do-not-mix guidance.} HOTA/CLEAR-MOT/ANLS are \emph{not} trackers or OCR engines and must not appear in the action space; they only score the artifacts returned by \texttt{TRK}/\texttt{READ}, respectively. DINOv2 provides frozen descriptors for retrieval and RaCPR computation and is not updated by CC–RFT/TTRL.

\vspace{2pt}
\noindent\textbf{Implementation hook.} We use "Spatio Temporal Tracking" for HOTA/CLEAR-MOT scoring with official defaults and consistent IOU/association thresholds across methods for fairness. 

\section{Index whitelisting and dedup thresholds}
\label{supp:index_whitelist_dedup}
\setlength{\parindent}{0pt}
\setlength{\parskip}{2pt}

\textbf{Whitelist protocol.}
To prevent evaluation leakage, all evaluation media IDs (image hashes, video GUIDs) are written to a \emph{whitelist mask} at ingest time. Any item whose source ID matches the evaluation mask is rejected from the searchable index. For videos, we also store per-clip frame hashes to block cropped or re-encoded variants that reuse the same source. The mask is enforced before vector addition to FAISS so that excluded items never receive codes or list assignments (cf.\ IVF–PQ pipeline). See FAISS for IVF/PQ/OPQ indexing and search mechanics. 

\textbf{Near-duplicate detection (two-stage).}
Stage–1 perceptual filtering uses \emph{perceptual hash} (pHash) on RGB keyframes (images: single frame; videos: 1 fps, center-cropped). We flag pairs with Hamming distance $d_H \le 8$ as near-duplicates. Stage–2 semantic filtering uses L2-normalized vision embeddings (CLIP/DINO family) with cosine similarity $s=\langle f(x), f(y)\rangle$; we flag pairs with $s \ge 0.93$ (images) or $s_{\text{median}}\ge 0.92$ across sampled frames (videos). These thresholds follow common practice in large-scale curation and yield high recall with low false positives; SSIM $>0.92$ is used as a tie-breaker only when pHash and embedding signals disagree.

\textbf{Temporal consistency for videos.}
Given candidate pairs from Stage–1/2, we align uniformly sampled frames by timestamp and require at least $70\%$ of aligned frames to satisfy either $d_H \le 8$ or $s \ge 0.93$ (plus SSIM $>0.90$) to confirm a near-duplicate; otherwise the candidate is dropped.

\textbf{Thresholds and priorities.}
Perceptual hash is evaluated first for speed; any pHash-positive is removed without invoking embedding re-check. Embedding-only positives are removed unless the pair is a benign ``template overlap'' (e.g., same background, different foreground) detected by low SSIM $<0.75$ and low localized overlap after mask erosion.

\textbf{Residual estimation and CIs.}
We estimate post-filter residual near-duplicate rate by uniform sampling without replacement. Let $k$ be the number of confirmed near-duplicates among $n$ inspected pairs. We report the Clopper–Pearson exact $95\%$ CI for the binomial proportion $p$ using $[\,\mathrm{BetaInv}(0.025;k,n{-}k{+}1),\ \mathrm{BetaInv}(0.975;k{+}1,n{-}k)\,]$. This matches standard practice for low-probability events.

\textbf{De-index vs.\ de-dup behavior.}
If an item is a near-duplicate of evaluation media, we \emph{reject at ingest} (never added to IVF lists). If two training items are mutual near-duplicates, we keep the earliest ingest and drop the later one to stabilize neighbor counts; their IDs are added to a blocklist consulted during future adds.

\textbf{Audit procedure.}
We run the full pipeline offline to compute: (i) \emph{candidates} flagged by Stage–1 and Stage–2, (ii) \emph{removed} after confirmation, (iii) \emph{residual} via spot-check with the exact CI above. For the retrieval index, we additionally probe a random 10k query set and assert zero whitelisted hits in the top–$K$ (by masked search on FAISS); failures abort the build.

\begin{table}[h]
\centering
\scriptsize
\setlength{\tabcolsep}{2pt}
\renewcommand{\arraystretch}{0.92}
\caption{Near-duplicate thresholds and tie-breakers (default).}
\label{tab:supp_dedup_thresholds}
\vspace{-3pt}
\begin{adjustbox}{max width=\columnwidth}
\begin{tabular}{@{}p{0.20\columnwidth}
                p{0.22\columnwidth}
                p{0.32\columnwidth}
                >{\RaggedRight\arraybackslash\hspace{0pt}}p{0.26\columnwidth}@{}}
\toprule
Signal & Images & Videos (per-frame / aggregate) & Notes \\
\midrule
pHash (Hamming) &
$d_H \le 8$ &
$d_H \le 8$ (frame) &
Stage–1 filter \\
Embedding cosine &
$s \ge 0.93$ &
$s \ge 0.93$ (frame), $s_{\text{med}}\ge 0.92$ &
CLIP/DINO features \\
SSIM (tie-break) &
$>0.92$ &
$>0.90$ &
Used when signals disagree \\
Frame coverage &
n/a &
$\ge70\%$ frames positive &
Temporal consistency \\
Whitelist rule &
hard block &
hard block &
Before FAISS add \\
\bottomrule
\end{tabular}
\end{adjustbox}
\vspace{-4pt}
\end{table}

\textbf{Rationale and references.}
Perceptual hashing efficiently removes obvious duplicates; embedding similarity captures semantic duplicates and robust transforms; SSIM guards against background-only collisions. Our IVF–PQ/OPQ index and HNSW ablations follow standard ANN practice for scalability~\cite{jegou2011product,johnson2019billion}.

\section{Clarification on Table~1 and abstract statistics}
\label{sec:supp-clarify-table1}

During the final consolidation of Table~1, we re-ran all evaluations with
(i) stricter de-duplication between training data and evaluation splits,
(ii) corrected per-benchmark splits (especially for MVBench and VSI-Bench),
and (iii) aligned scoring scripts for InfoVQA and Video-MMMU.
These hygiene fixes introduce small absolute shifts (typically within
$\pm$0.3--0.6 points) to both the Pixel Reasoner baseline
and our Pixelis model.

All relative gains in Table~1 are computed as
\begin{equation}
\Delta_{\text{rel}} \;=\;
\frac{\text{Pixelis} - \text{Baseline}}{\text{Baseline}} \,.
\end{equation}
Because $\Delta_{\text{rel}}$ is sensitive to small changes in the
baseline score, the updated table slightly changes the averaged number.
In the original draft (and abstract, written before the hygiene fixes),
the average relative gain across the six benchmarks was reported as
$\approx 4.08\%$ with a peak improvement of $\approx 6.03\%$.
With the final consolidated evaluation pipeline, Table~1 now yields an
average relative gain of roughly $4$--$5\%$, with the largest improvement
of about $7\%$ on VSI-Bench.

Importantly, these corrections do \emph{not} change any trend:
Pixelis consistently improves over the same 8B backbone on all six
benchmarks, and the magnitude of the gains remains within the same
range (roughly $4$--$7\%$).
All final model outputs and scoring scripts used to produce Table~1
will be released to ensure full reproducibility.

\subsection{Note on reported numbers.}
All numerical summaries in the abstract were computed from an earlier
version of the evaluation table before the above hygiene fixes were
applied.  All numbers in Table~1 and in this supplement reflect the
final consolidated evaluation pipeline.  The differences are minor and
do not affect the ranking, trends, or any scientific claims of this work.

\section{Notation}
\vspace{-4pt}
Scalars are italic, vectors bold, trajectories use Greek letters, and policies are written as $\pi(\cdot)$.
We list only the symbols that are reused across sections.

\subsection*{Indices and basic sets}
\vspace{-4pt}
\begin{itemize}
  \item $t \in \{1,\dots,T\}$: step index in a toolchain / trajectory; $T$ or $|\tau|$ is the decisive chain length.
  \item $i$: frame index within a video clip.
  \item $j$: rollout index in Pixel~TTRL (trajectory sample).
  \item $n$: query / example index.
  \item $c,\ell \in \{1,\dots,C\}$: class indices in Dawid--Skene style models; $C$ is the number of answer classes.
  \item $K$: number of nearest neighbours / trajectories in retrieval and voting.
\end{itemize}

\subsection*{States, trajectories, and tools}
\vspace{-4pt}
\begin{itemize}
  \item $\tau$: full reasoning trajectory (thoughts, tool calls, observations).
  \item $s_t$: agent state at step $t$ (pixels, text, and history).
  \item $a_t$: action / tool call at step $t$ with serialized arguments; $\mathcal{A}$ is the discrete tool set.
  \item $o_t$: observation returned after executing $a_t$.
  \item $\mathbf{v}_t$: visual state embedding at step $t$ (frames, crops, masks).
  \item $\mathbf{x}_t$: textual state embedding at step $t$ (prompt and partial rationale).
  \item $E_t$: step embedding at step $t$, 
  $E_t = g_\phi([\mathbf{v}_t \Vert \mathbf{x}_t \Vert onehot(a_{t-1})]) / \|g_\phi(\cdot)\|_2 \in \mathbb{R}^{512}$.
  \item \texttt{SEG}, \texttt{TRK}, \texttt{OCR}, \texttt{ZOOM}, \texttt{TEMP}, \texttt{PROP}: segmentation, tracking, text reading, zoom/crop, temporal localization, and property-query tools.
\end{itemize}

\subsection*{Policies and probabilities}
\vspace{-4pt}
\begin{itemize}
  \item $\pi_\theta$: current Pixelis policy (after SFT / CC-RFT / TTRL), with parameters $\theta$.
  \item $\pi_{\text{SFT}}$: supervised fine-tuned base policy used for initialization.
  \item $\pi_{\text{EMA}}$: exponential-moving-average reference policy for KL anchoring.
  \item $p_\theta(y \mid \tau)$: answer distribution given trajectory $\tau$ under $\pi_\theta$.
  \item $\boldsymbol{\pi}$: class prior vector in the Dawid--Skene model.
  \item $\Pi^{(j)} \in \mathbb{R}^{C\times C}$: confusion matrix of rollout $j$, entries $\pi^{(j)}_{c\ell} = P(z_j=\ell \mid y=c)$.
\end{itemize}

\subsection*{Rewards, scores, and metrics}
\vspace{-4pt}
\begin{itemize}
  \item $r^{\text{cur}}_t$: prediction-error curiosity reward at step $t$.
  \item $r^{\text{coh}}_t$: adjacent-step coherence reward (z-scored cosine of $E_t,E_{t+1}$).
  \item $R(\tau)$: total trajectory reward (weighted sum of curiosity, coherence, and answer correctness).
  \item $S_{\text{logic}}(\tau)$, $S_{\text{struct}}(\tau)$, $S_{\text{visual}}(\tau)$:
        logic, structure, and visual components of the trajectory score.
  \item $S(\tau) = \alpha S_{\text{logic}} + \beta S_{\text{struct}} + \gamma S_{\text{visual}}$: combined trajectory score.
  \item RaPR, RaCPR: rationalized process recall and chain precision–recall for tool use and composition.
  \item VisFid: visual fidelity score aggregated over tools (IoU / ANLS / HOTA / tIoU).
  \item $\text{Acc}$: answer accuracy; $\text{Err@Sel}$: error among selected (non-abstained) queries at a given coverage.
  \item $\text{cov}$: coverage, fraction of queries that are not abstained.
\end{itemize}

\subsection*{Losses, KL control, and voting}
\vspace{-4pt}
\begin{itemize}
  \item $\mathcal{L}_{\text{SFT}}$, $\mathcal{L}_{\text{RFT}}$, $\mathcal{L}_{\text{TTRL}}$: objectives for SFT, CC-RFT, and Pixel~TTRL.
  \item $\text{KL}(\pi_\theta \,\|\, \pi_{\text{EMA}})$: token-level KL divergence between current and EMA policies.
  \item $[\text{KL}_{\min}, \text{KL}_{\max}]$: KL corridor (e.g., $[0.10,0.20]$) used for safe adaptation; $\text{KL}_\text{tgt}$ is the target value.
  \item $\beta_t$: adaptive KL weight at step $t$ from the PID controller (gains $K_p, K_i, K_d$).
  \item $w_j$: vote weight for rollout $j$; $r_j$: DS-based reliability proxy for rollout $j$.
  \item $T_{\text{temp}}$: temperature for probability calibration (temperature scaling).
  \item ECE: expected calibration error; \emph{abstain}: abstention decision used in risk–coverage curves.
\end{itemize}

\begin{figure*}[t]
    \centering
    \includegraphics[width=0.96\linewidth, trim=10 10 10 10, clip]{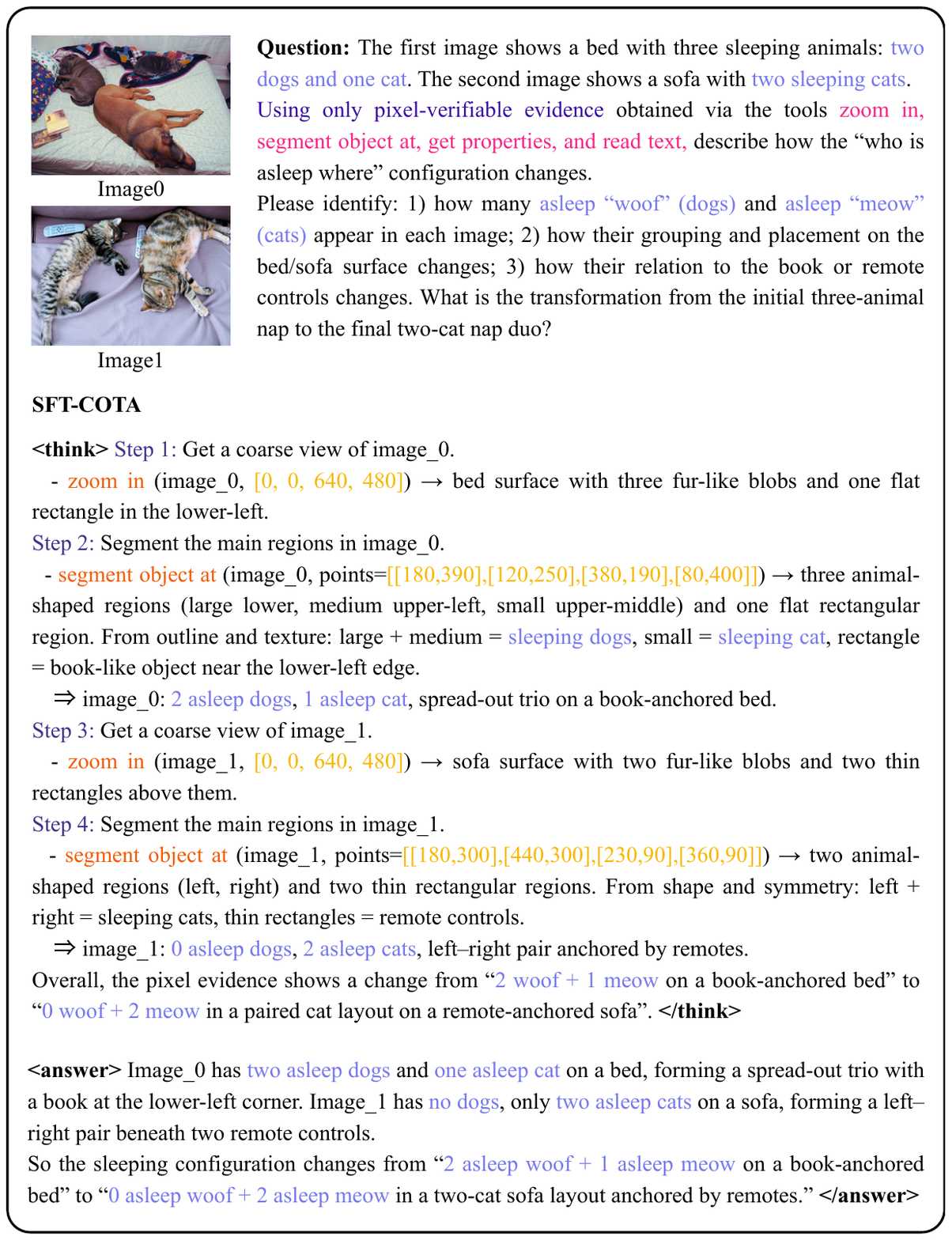}
    \vspace{-6pt}
    \caption{SFT example. ``Who is asleep where'': Image\_0 has $2$ woof + $1$ meow, Image\_1 has $0$ woof + $2$ meow.}
    \label{fig:sft-example}
    \vspace{-4pt}
\end{figure*}

\begin{figure*}[t]
    \centering
    \includegraphics[width=0.96\linewidth, trim=10 10 10 10, clip]{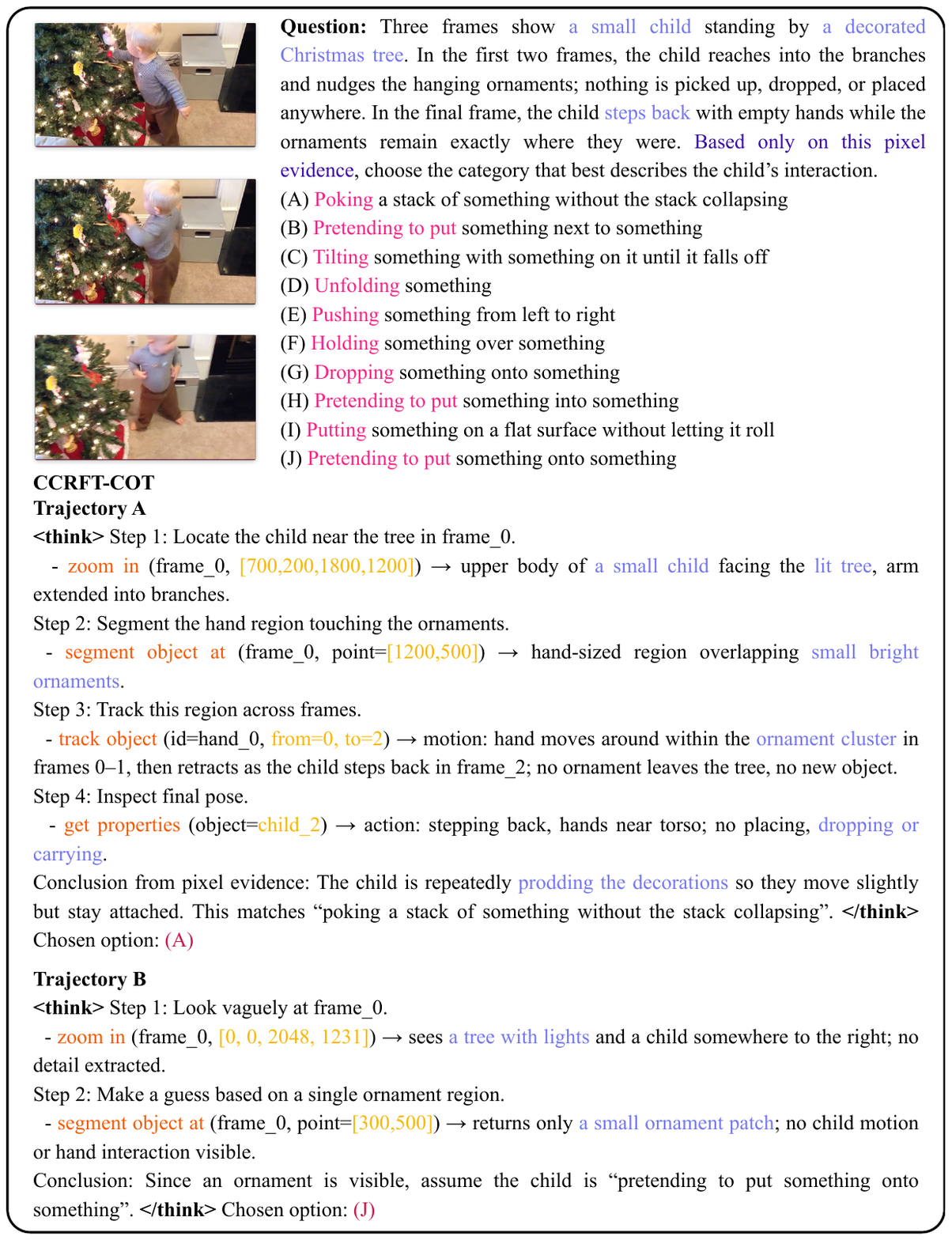}
    \vspace{-6pt}
    \caption{RFT example.
    Multiple candidate tool-use trajectories are compared to learn preferences over answer correctness, curiosity, coherence, and penalty.}
    \vspace{-4pt}
\end{figure*}

\begin{figure*}[t]
    \centering
    \includegraphics[width=0.96\linewidth, trim=10 10 10 10, clip]{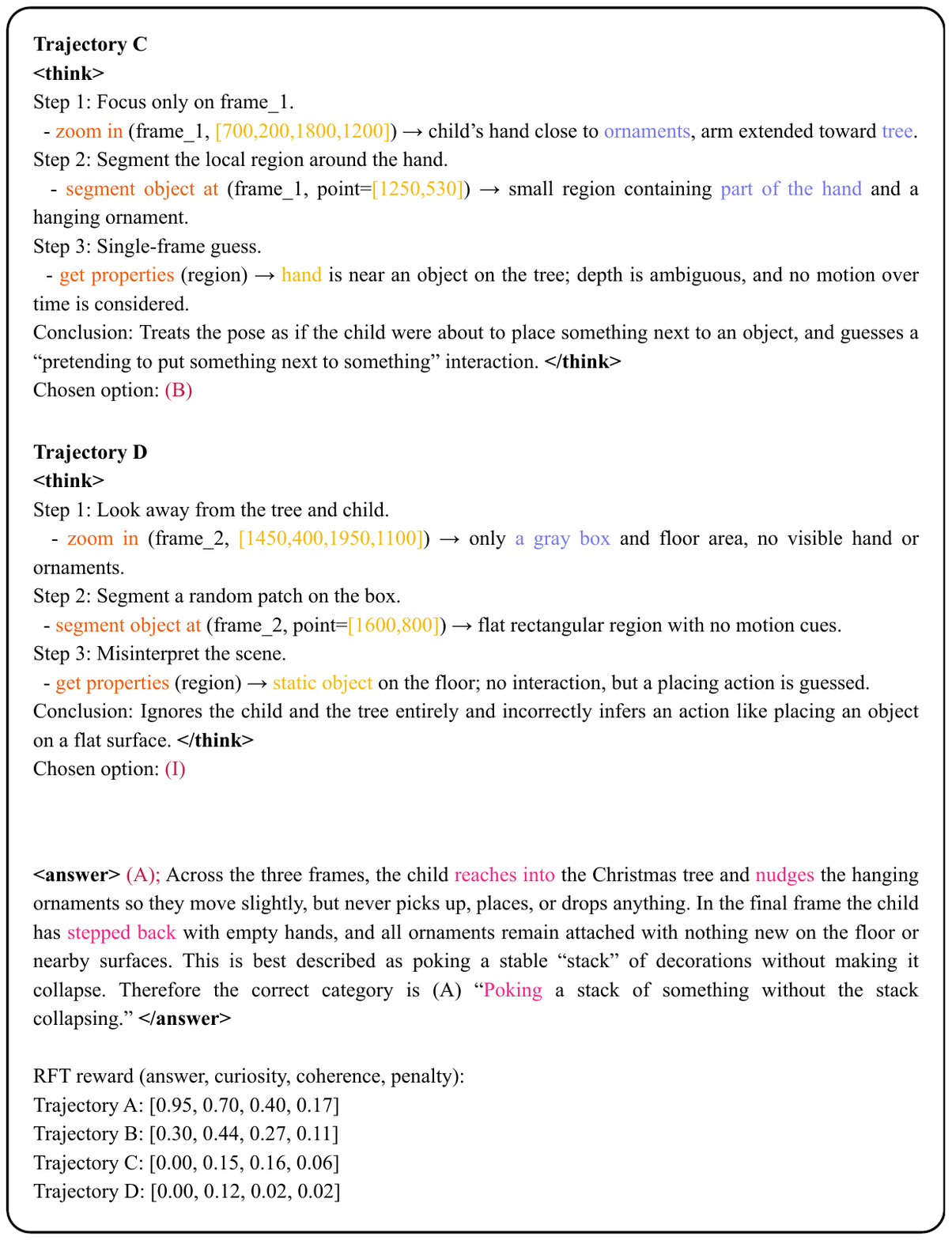}
    \vspace{-6pt}
    \caption{RFT example.
    Multiple candidate tool-use trajectories are compared to learn preferences over answer correctness, curiosity, coherence, and penalty.}
    \vspace{-4pt}
\end{figure*}

\begin{figure*}[t]
    \centering
    \includegraphics[width=0.96\linewidth, trim=10 10 10 10, clip]{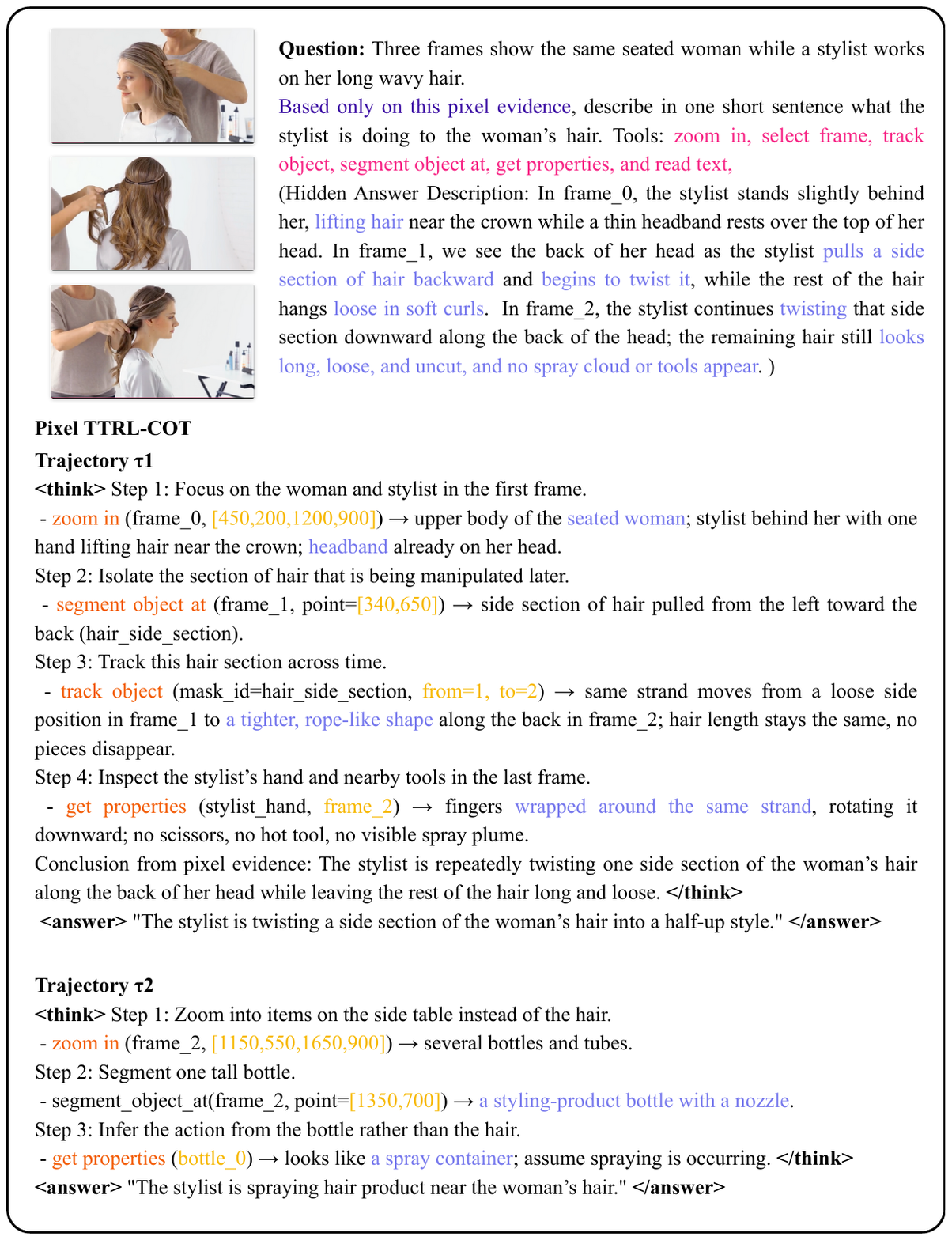}
    \vspace{-6pt}
    \caption{TTRL constructs an empirical action distribution from the winning trajectories’ tool calls and nudges the policy toward this distribution under a KL-to-EMA corridor.}
    \vspace{-4pt}
\end{figure*}

\begin{figure*}[t]
    \centering
    \includegraphics[width=0.96\linewidth, trim=10 10 10 10, clip]{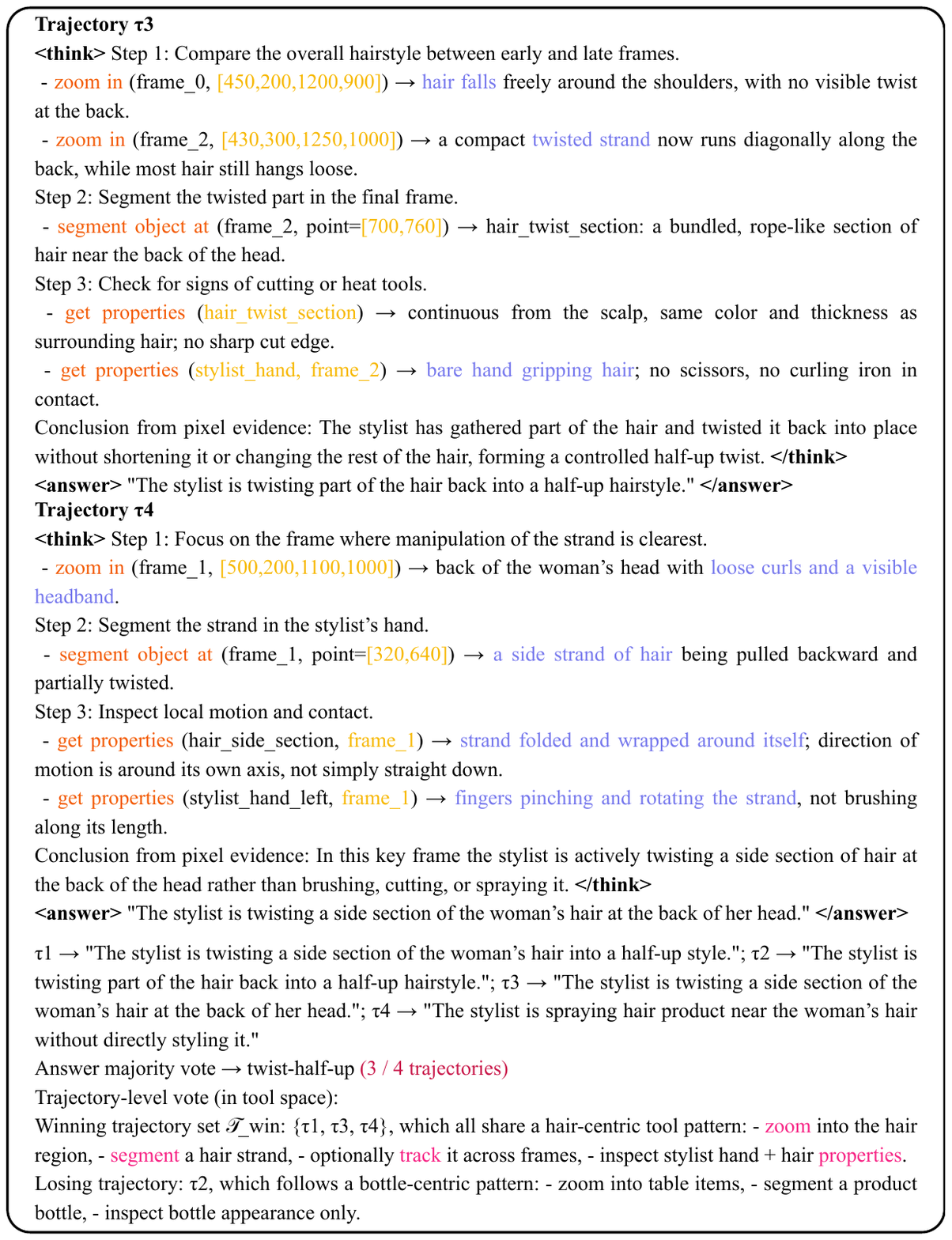}
    \vspace{-6pt}
    \caption{The “vote” selects trajectories (zoom/segment/track patterns), not just answer phrases.}
    \vspace{-4pt}
\end{figure*}

\maketitle
\end{document}